# Proximity-Based Non-uniform Abstractions for Approximate Planning

**Jiří Baum**  Jiri@baum.com.au
**Ann E. Nicholson**  Ann.Nicholson@monash.edu
**Trevor I. Dix**  Trevor.Dix@monash.edu
*Faculty of Information Technology*
*Monash University, Clayton, Victoria, Australia*

## Abstract

In a deterministic world, a planning agent can be certain of the consequences of its planned sequence of actions. Not so, however, in dynamic, stochastic domains where Markov decision processes are commonly used. Unfortunately these suffer from the 'curse of dimensionality': if the state space is a Cartesian product of many small sets ('dimensions'), planning is exponential in the number of those dimensions.

Our new technique exploits the intuitive strategy of selectively ignoring various dimensions in different parts of the state space. The resulting non-uniformity has strong implications, since the approximation is no longer Markovian, requiring the use of a modified planner. We also use a spatial and temporal proximity measure, which responds to continued planning as well as movement of the agent through the state space, to dynamically adapt the abstraction as planning progresses.

We present qualitative and quantitative results across a range of experimental domains showing that an agent exploiting this novel approximation method successfully finds solutions to the planning problem using much less than the full state space. We assess and analyse the features of domains which our method can exploit.

## 1. Introduction

In a deterministic world where a planning agent can be certain of the consequences of its actions, it can plan a sequence of actions, knowing that their execution will necessarily achieve its goals. This assumption is not appropriate for flexible, multi-purpose robots and other intelligent software agents which need to be able to plan in the dynamic, stochastic domains in which they will operate, where the outcome of taking an action is uncertain.

For small to medium-sized stochastic domains, the theory of Markov decision processes provides algorithms for generating the optimal plan (Bellman, 1957; Howard, 1960; Puterman & Shin, 1978). This plan takes into account uncertainty about the outcome of taking an action, which is specified as a distribution over the possible outcomes. For flexibility, there is a reward function rather than a simple goal, so that the relative desirability or otherwise of each situation can be specified.

However, as the domain becomes larger, these algorithms become intractable and approximate solutions become necessary (for instance Drummond & Bresina, 1990; Dean, Kaelbling, Kirman, & Nicholson, 1995; Kim, 2001; Steinkraus, 2005). In particular where the state space is expressed in terms of dimensions, or as a Cartesian product of sets, its size and the resulting computational cost is exponential in the number of dimensions. On the other hand, fortunately, this results in a fairly structured state-space where effective approximations often should be possible.

Our solution is based on selectively ignoring some of the dimensions, in some parts of the state space, some of the time. In other words, we obtain approximate solutions by dynamically varying the level of abstraction in different parts of the state space. There are two aspects to this approach. Firstly, the varying level of abstraction introduces some artefacts, and the planning algorithm must be somewhat modified so as to eliminate these. Secondly, more interestingly, an appropriate abstraction must be selected and later modified as planning and action progress.

Our work is an extension and synthesis of two existing approaches to approximate planning: the locality-based approximation of envelope methods (Dean et al., 1995) and the structure-based approximation of uniform abstraction (Nicholson & Kaelbling, 1994; Dearden & Boutilier, 1997). Our work extends both of these by exploiting both structure and locality, broadening the scope of problems that can be contemplated. Baum and Nicholson (1998) introduced the main concepts while full details of our algorithms and experimental results are presented in Baum's (2006) thesis. There have been some studies of arbitrary abstraction, for instance by Bertsekas and Tsitsiklis (1996). However, these are generally theoretical and in any case they tended to treat the approximation as Markovian, which would have resulted in unacceptable performance in practice. We improve on this by extending the planning algorithm to deal with the non-Markovian aspects of the approximation. Finally, we use a measure of locality, introduced by Baum and Nicholson (1998), that is similar to but more flexible than the influence measure of Munos and Moore (1999).

We assume that the agent continues to improve its plan while it is acting and that planning failures are generally not fatal. We also deal with control error exclusively. Sensor error is not considered and it is assumed that the agent can accurately discern the current world state ("fully observable"), and that it accurately knows the state space, the goal or reward function, and a distribution over the effect of its actions (no learning).

The remainder of this paper is organised as follows. Section 2 reviews the background, introduces our abstraction and provides our framework. Section 3 discusses planning under a static non-uniform abstraction, and Section 4 presents our method for initially selecting the non-uniform abstraction based on the problem description. Section 5 presents a method of changing the abstraction based on the policy planned, while Sections 6 and 7 introduce a proximity measure and a method of varying the abstraction based on that measure, respectively. Section 8 presents results based both on direct evaluation of the calculated policy and on simulation. Finally, Section 9 discusses the results and Section 10 gives our conclusions and outlines possible directions for future work.

## 2. Planning under Non-uniform Abstractions

In a non-deterministic world where a planning agent cannot be certain of the consequences of its actions except as probabilities, it cannot plan a simple sequence of actions to achieve its goals. To be able to plan in such dynamic, stochastic domains, it must use a more sophisticated approach. Markov decision processes are an appropriate and commonly used representation for this sort of planning problem.

## 2.1 Illustrative Problems

To aid in exposition, we present two example problems here. The full set of experimental domains is presented in Section 8.1.

The two illustrative problems are both from a grid navigation domain, shown in Figure 1. They both have integer $x$ and $y$ coordinates from 0 to 9, three doors which can be either open or closed and a damage indication which can be either yes or no. The agent can move in one of the four cardinal directions, open a door if it is next to it, or do nothing. The doors are fairly difficult to open, with probability of success 10% per time step, while moving has an 80% chance of success, with no effect in the case of failure. Running into a wall or into a closed door causes damage, which cannot be repaired. The transitions are shown in Table 1. The agent starts in the location marked $s^0$ in Figure 1 with the doors closed and no damage, and the goal is to reach the location marked $*$ with no damage.

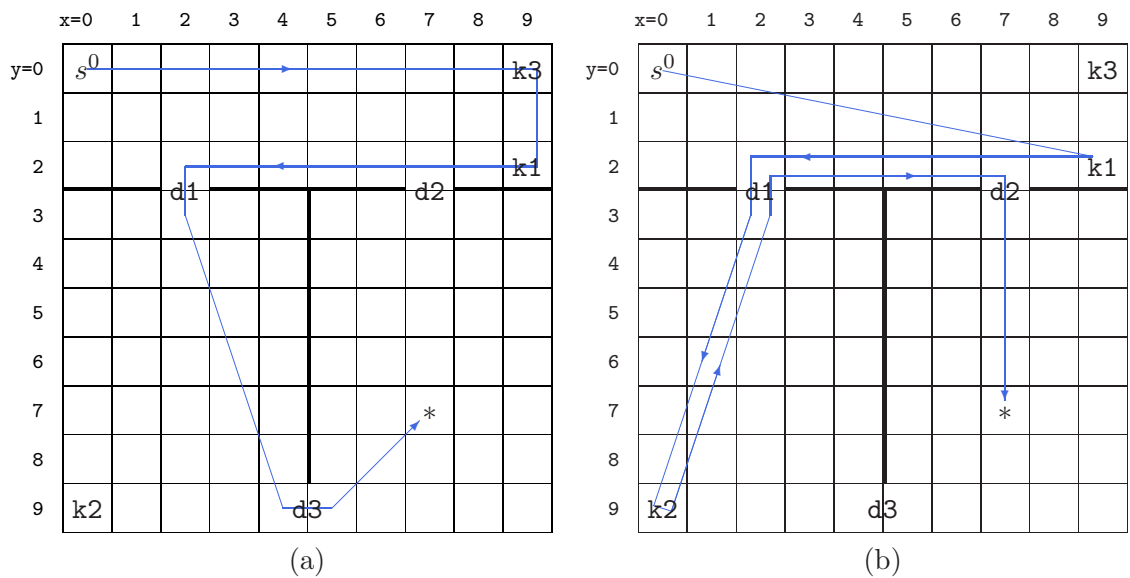

Figure 1: The layout of the grid navigation domain. The blue arrows show the optimal path (a) and a suboptimal path (b) through the 3Keys problem. The 3Doors problem has the same grid layout (walls and doors) but with no keys.

The 3Keys problem also contains keys, which are required to open the doors. The agent may have any one or more of them at any time. An additional action allows the agent to pick up each key in the location as shown in the figure, and the open action requires the corresponding key to be effective (there is no separate 'unlock' action). The 3Doors problem contains no keys — the doors are unlocked but closed — and therefore no corresponding keys or pickup action.

The optimal policy obtained by exact planning for the 3Doors problem simply takes the shortest path through door 2. For the 3Keys problem, the optimal plan is to collect keys 3 and 1, pass south through door 1 and east through door 3, shown in Figure 1(a). A suboptimal plan is shown in Figure 1(b).

|  | pre-state | | | | |  | | post-state | | | | | |
| --- | --- | --- | --- | --- | --- | --- | --- | --- | --- | --- | --- | --- | --- |
| x | y | d1 | d2 | d3 | dmg | → | | x | y | d1 | d2 | d3 | dmg |
| **Stay** | | | | | | | | | | | | | |
| . | . | . | . | . | . | → | | . | . | . | . | . | . |
| **South** | | | | | | | | | | | | | |
| 2 | 2 | open | . | . | . | → | 80% | . | 3 | . | . | . | . |
| 7 | 2 | . | open | . | . | → | 80% | . | 3 | . | . | . | . |
| . | 2 | . | . | . | . | → | | . | . | . | . | . | yes |
| . | 9 | . | . | . | . | → | | . | . | . | . | . | yes |
| . | $y$ | . | . | . | . | → | 80% | . | $y+1$ | . | . | . | . |
| **North** | | | | | | | | | | | | | |
| 2 | 3 | open | . | . | . | → | 80% | . | 2 | . | . | . | . |
| 7 | 3 | . | open | . | . | → | 80% | . | 2 | . | . | . | . |
| . | 0 | . | . | . | . | → | | . | . | . | . | . | yes |
| . | 3 | . | . | . | . | → | | . | . | . | . | . | yes |
| . | $y$ | . | . | . | . | → | 80% | . | $y-1$ | . | . | . | . |
| **East** | | | | | | | | | | | | | |
| 4 | 0 | . | . | . | . | → | 80% | 5 | . | . | . | . | . |
| 4 | 1 | . | . | . | . | → | 80% | 5 | . | . | . | . | . |
| 4 | 2 | . | . | . | . | → | 80% | 5 | . | . | . | . | . |
| 4 | 9 | . | . | open | . | → | 80% | 5 | . | . | . | . | . |
| 4 | . | . | . | . | . | → | | . | . | . | . | . | yes |
| 9 | . | . | . | . | . | → | | . | . | . | . | . | yes |
| $x$ | . | . | . | . | . | → | 80% | $x+1$ | . | . | . | . | . |
| **West** | | | | | | | | | | | | | |
| 5 | 0 | . | . | . | . | → | 80% | 4 | . | . | . | . | . |
| 5 | 1 | . | . | . | . | → | 80% | 4 | . | . | . | . | . |
| 5 | 2 | . | . | . | . | → | 80% | 4 | . | . | . | . | . |
| 5 | 9 | . | . | open | . | → | 80% | 4 | . | . | . | . | . |
| 5 | . | . | . | . | . | → | | . | . | . | . | . | yes |
| 0 | . | . | . | . | . | → | | . | . | . | . | . | yes |
| $x$ | . | . | . | . | . | → | 80% | $x-1$ | . | . | . | . | . |
| **Open** | | | | | | | | | | | | | |
| 2 | 2 | . | . | . | . | → | 10% | . | . | open | . | . | . |
| 2 | 3 | . | . | . | . | → | 10% | . | . | open | . | . | . |
| 7 | 2 | . | . | . | . | → | 10% | . | . | . | open | . | . |
| 7 | 3 | . | . | . | . | → | 10% | . | . | . | open | . | . |
| 4 | 9 | . | . | . | . | → | 10% | . | . | . | . | open | . |
| 5 | 9 | . | . | . | . | → | 10% | . | . | . | . | open | . |
| . | . | . | . | . | . | → | | . | . | . | . | . | yes |

Table 1: Transitions in the `3Doors` problem, showing important and changed dimensions only. First matching transition is used. Where a percentage is shown, the given post-state will occur with that probability, otherwise the state is unchanged. Transitions without percentages are deterministic.

## 2.2 Exact Planning

One approach to exact planning in stochastic domains involves using Markov Decision Processes (MDPs). A MDP is a tuple $\langle \mathcal{S}, A, T, R, s^0 \rangle$, where $\mathcal{S}$ is the *state space*, $A$ is the *set of* available *actions*, $T$ is the *transition function*, $R$ is the *reward function* and $s^0 \in S$ is the *initial state*. The agent begins in state $s^0$. At each time step, the agent selects an action $a \in A$, which, together with the current state, applies $T$ to obtain a distribution over $\mathcal{S}$. The current state at the next time-step is random according to this distribution and we write $\Pr_T(s, a, s')$ for the probability that action $a$ taken in state $s$ will result in the state $s'$ at the next time-step. The agent is also given a reward at each time step, calculated by $R$ from the current state (and possibly also the action selected). The aim of the agent is to maximise some cumulative function of these rewards, typically the expected discounted[1] sum under a discounting factor $\gamma$. In a fully-observable MDP, the agent has full knowledge. In particular, the agent is aware of $T$, $R$ and the current state when selecting an action. It is well-known that in a fully-observable MDP, the optimal solution can be expressed as a policy $\pi : \mathcal{S} \to A$ mapping from the current state to the optimum action. The planning problem, then, is the calculation of this $\pi$. As a side-effect of the calculation, the standard algorithms also calculate a value function $V : \mathcal{S} \to \mathbb{R}$, the expected discounted sum of rewards starting from each state. Table 2 summarises the notation used in this paper.

There are well known iterative algorithms, Bellman's (1957) value iteration, Howard's (1960) policy iteration, and the modified policy iteration of Puterman and Shin (1978) for computing the optimal policy $\pi^*$. However, as $\mathcal{S}$ becomes larger, the calculation of $\pi^*$ becomes more computationally expensive. This is particularly so if state space is structured as a Cartesian product of *dimensions*, $\mathcal{S} = \mathcal{S}_1 \times \mathcal{S}_2 \times \cdots \times \mathcal{S}_D$, because then $|\mathcal{S}|$ is exponential in $D$. Since the algorithms explicitly store $V$ and usually $\pi$, which are functions of $\mathcal{S}$, their space complexity is therefore also exponential in $D$. Since they iterate over these arrays, the time complexity is also at least exponential in $D$, even before any consideration of how fast these iterative algorithms themselves converge. Typically, as $D$ grows, planning in $\mathcal{S}$ quickly becomes intractable. Since in practice the amount of computation allowed to the agent is limited, this necessitates some approximations in the process.

In the `3Doors` problem, there are six dimensions (two for the $x$ and $y$ coordinates, three for the doors and one for damage), so that $\mathcal{S} = \{0 \ldots 9\} \times \{0 \ldots 9\} \times \{\text{open}, \text{closed}\} \times \{\text{open}, \text{closed}\} \times \{\text{open}, \text{closed}\} \times \{\text{damage}, \text{no damage}\}$ and $|\mathcal{S}| = 12\,800$. The action space $A$ is a set of five actions, $A = \{\text{north}, \text{south}, \text{east}, \text{west}, \text{open}\}$. The transition function specifies the outcomes of taking each action in each state. The reward function $R$ is 0 in the agent is in the $\langle 7, 7 \rangle$ location (marked $*$ in the diagram) with no damage, $-1$ if it is in any other location with no damage, and $-2$ if there is damage. Finally, $s^0$ is the state where the agent is in the $\langle 0, 0 \rangle$ location, the doors are closed, and there is no damage.

Exact planning is listed in our results with $|\mathcal{S}|$ and $V^*(s^0)$ for comparison. If there is no approximation, the planner must consider the whole state space $\mathcal{S}$. $|\mathcal{S}|$ is therefore a measure of the cost of this planning — directly in terms of space and indirectly in terms of time. On the other hand, since the planning is exact, the optimal value function $V^*$ will be obtained, along with the optimal policy $\pi^*$ ensuring that the agent will expect to obtain that value. It is against these figures that all approximations must measure.

---

[1]. While our illustrative problems have simple goals of achievement, we use time discounting in order to remain general and for its mathematical convenience.

|  | symbol |  | meaning |
|---|---|---|---|
| original | | abstract | |
| $\mathcal{S}$ | | $\mathcal{W} \subset \mathcal{P}(\mathcal{S})$ | state space (specific state space / worldview, resp.) |
| $\mathcal{S}_d$ | | — | dimension $d$ of the state space |
| | $D \in \mathbb{N}^*$ | | number of dimensions |
| $s \in \mathcal{S}$ | | $w \in \mathcal{W}$ | a state |
| $s^0 \in \mathcal{S}$ | | — | initial state |
| $s^{\text{cur}} \in \mathcal{S}$ | | — | current state (in on-line planning) |
| $s_d \in \mathcal{S}_d$ | | $w_d \subseteq \mathcal{S}_d$ | dimension $d$ of the state $s$ or $w$, resp. |
| | $A$ | | set of actions (action space) |
| | $a_0 \in A$ | | default action |
| $T$ | | $T$ | transition function (formal) |
| $\Pr_T:\mathcal{S}\times A\times\mathcal{S}\to[0,1]$ | | $\Pr_T:\mathcal{W}\times A\times\mathcal{W}\to[0,1]$ | transition function (in use) |
| $R:\mathcal{S}\to\mathbb{R}$ | | $R:\mathcal{W}\to\mathbb{R}$ | reward function (one-step reward) |
| $V:\mathcal{S}\to\mathbb{R}$ | | $V:\mathcal{W}\to\mathbb{R}$ | value function (expected discounted sum of rewards) |
| | $\gamma \in [0,1)$ | | discount factor for the reward |
| $\pi:\mathcal{S}\to A$ | | $\pi:\mathcal{W}\to A$ | policy |
| $\pi^*:\mathcal{S}\to A$ | | — | optimal policy |
| $V^*:\mathcal{S}\to\mathbb{R}$ | | — | optimal value function (exact value function of $\pi^*$) |
| $\hat{\pi},\hat{\pi}_i:\mathcal{S}\to A$ | | $\hat{\pi},\hat{\pi}_i:\mathcal{W}\to A$ | approximate policy, $i$th approximate policy |
| $V^*_{\hat{\pi}}:\mathcal{S}\to\mathbb{R}$ | | — | exact value function of $\hat{\pi}$ (note: $\hat{\pi}$ may be abstract) |
| $\hat{V}_{\hat{\pi}}:\mathcal{S}\to\mathbb{R}$ | | $\hat{V}_{\hat{\pi}}:\mathcal{W}\to\mathbb{R}$ | approximate value function of $\hat{\pi}$ (approx. to $V^*_{\hat{\pi}}$) |

Table 2: Summary of notation. The first column is the notation for the original MDP, the second is the notation once non-uniform abstraction has been applied.

### 2.3 Uniform Abstraction

One method of approximation is to take advantage of these dimensions by ignoring some of them — those that are irrelevant or only marginally relevant — in order to obtain an approximate solution. It is uniform in the sense that the same dimensions are ignored throughout the state space. Since this approach attacks the curse of dimensionality where it originates, at the dimensions, it should be very effective at counteracting it.

Dearden and Boutilier use this to obtain an exact solution (Boutilier, 1997) or an approximate one (Boutilier & Dearden, 1996; Dearden & Boutilier, 1997). However, their abstractions are fixed throughout execution, and dimensions are also deleted from the problem in a pre-determined sequence. This makes their approach somewhat inflexible. Similarly, Nicholson and Kaelbling (1994) propose this technique for approximate planning. They delete dimensions from the problem based on sensitivity analysis, and refine their abstraction as execution time permits, but it is still uniform. Dietterich (2000) uses this kind of abstraction in combination with hierarchical planning to good effect: a subtask, such as 'Navigate to location $t$', can ignore irrelevant dimensions, such as location of items to be picked up or even the ultimate destination of the agent. Generally, any time the problem description is derived from a more general source rather than specified for the particular problem, uniform abstraction will help. Gardiol and Kaelbling (2008) use it where some dimensions are relevant, but only marginally, so that ignoring them results in an approximate solution that can be improved as planning progresses.

Unfortunately, however, at least with human-specified problems, one would generally expect all or most mentioned dimensions to be in some way relevant. Irrelevant dimensions will be eliminated by the human designer in the natural course of specifying the problem. Depending on the domain and the situation some marginally relevant dimensions might be included, but often, these will not be nearly enough for effective approximation.

We do not list comparisons against uniform abstraction in our results for this reason — in most of our sample domains, it makes little sense. All or almost all of the dimensions are important to solving the problem. Where this is not the case and methods exist for effective uniform abstraction, they can be integrated with our approach easily.

### 2.4 Non-uniform Abstraction

Our approximation, *non-uniform abstraction*, replaces the state space $\mathcal{S}$ with $\mathcal{W}$, a particular type of partition of $\mathcal{S}$, as originally introduced by Baum and Nicholson (1998). We call $\mathcal{W}$ the *worldview*, so members of $\mathcal{W}$ are *worldview states* while members of $\mathcal{S}$ are *specific states*.[2] Non-uniform abstraction is based on the intuitive idea of ignoring some dimensions in *some parts* of the state space. For example, a door is only of interest when the agent is about to walk through it, and can be ignored in those parts of the state space which are distant from the door. In a particular member of the worldview $w^i \in \mathcal{W}$, each dimension is either taken into account completely (*concrete*, refined in), or ignored altogether (*abstract*, coarsened out). $w^i = \prod_{d=1}^{D} w_d^i$ and each $w_d^i$ is a singleton subset of the corresponding $\mathcal{S}_d$ for concrete dimensions and equal to $\mathcal{S}_d$ for abstract dimensions.[3] It is up to the worldview selection and modification methods to ensure that $\mathcal{W}$ remains a partition at all times.

To give an example, in the `3Doors` problem one possible worldview has the location and damage dimensions concrete in every state, while the door dimensions are each concrete only in states within two steps of the respective door.

Note that the domain is still fully-observable. This is not a question of lack of knowledge about the dimensions in question, but wilful conditional ignorance of them during planning as a matter of computational expediency. The approximation also subsumes both exact planning and uniform abstraction. For exact planning, all dimensions can be set uniformly concrete, so that $|\mathcal{W}| = |\mathcal{S}|$ and each worldview state corresponds to one specific state. For uniform abstraction, the combination of abstract and concrete dimensions can be fixed for the entire worldview. They can be treated as special cases of our more general approach.[4]

---

2. Previously, we used the word 'envelope' for the same concept (Baum & Nicholson, 1998), however, 'worldview' better describes the approximation used than 'envelope'.
3. We do not allow a dimension to be partially considered, we only abstract to the level of dimensions, not within them. A dimension such as the $x$ coordinate will either have a particular value, or it will be fully abstract, but it will never be 5–9, for instance.
4. Our modified $\pi$ calculation reduces to the standard algorithm for uniform or fully concrete worldviews, so our planner obtains the standard results in these cases.

On the other hand, the approximation is no longer Markovian. A dimension that is abstracted away is indeterminate. In the notation of Markov Decision Processes, this can only be represented by some distribution over the concrete states, but the dimension is not stochastic — it has some specific (but ignored) value. The distinction is important because for a truly stochastic outcome, it can be quite valid to plan to retry some action until it 'succeeds' (for instance, opening a door in the 3Doors problem). For a dimension which is merely ignored, the agent will obtain the same outcome (door is closed) each time it moves into the region where the dimension is not ignored, so that within the worldview, previous states can appear to matter. We discuss this further in Section 3.

### 2.5 Comparison to Other Approaches

Non-uniform abstractions began to appear in the literature at first usually as a side-effect of a structured method, where the state space is represented as a decision tree based on the individual dimensions, such as Boutilier, Dearden, and Goldszmidt (1995, 2000). Note, however, that the decision tree structure imposes a restriction on the kinds of non-uniform abstraction that can be represented: the dimension at the root of the tree is considered throughout the state space, and so on. This is a significant restriction and results in a representation much more limited than our representation. A similar restriction affects de Alfaro and Roy's (2007) "magnifying-lens abstraction", with the refinement that multi-valued dimensions are taken bit-by-bit and the bits interleaved, so that each level of the decision tree halves the space along a different dimension in a pre-determined order. As they note, this would work well where these dimensions correspond to a more-or-less connected space, as in a gridworld, but it would do less well with features like the doors of our grid navigation domain. Magnifying-lens abstraction calculates upper and lower bounds to the value function, rather than a single approximation, which is an advantage for guiding abstraction selection and allows for a definite termination condition (which we lack). On the other hand, it always considers fully-concrete states in part of the algorithm, limiting its space savings to the square root of the state space, whereas our algorithm can work with a mixture of variously abstract states not necessarily including any fully concrete ones. Another related approach is variable grids used for discretisation, which can be indirectly used for some discrete domains, as Boutilier, Goldszmidt, and Sabata (1999) do, if dimensions can be reasonably approximated as continuous (for instance money). Unlike our approach, variable grids are completely inapplicable to predicates and other binary or enumerated dimensions. Some, such as Reyes, Sucar, and Morales (2009), use techniques in some ways quite similar to ours for continuous MDPs, though they are quite different in other ways: they consider refinement only, not coarsening; they use sampling, rather than directly dealing with the domain model; and they use a different refinement method, where each refinement is evaluated after the fact and then either committed or rolled back.

Perhaps the most similar to our approach is one of the modules of Steinkraus (2005), the ignore-state-variables module. However, the module appears to be completely manual, requiring input of which variables (dimensions) should be ignored in what parts of the state space. It also uses the values of the dimensions from the current state $s^{\text{cur}}$, rather than a distribution, which obviously restricts the situations in which it may be used (for instance, in the 3Doors problem, the doors could not be ignored in the starting state). Finally, since Steinkraus (2005) does not analyse or report the relative contributions of the modules to the solution, nor on the meta-planning problem of selecting and arranging the modules, it is difficult to know to what extent this particular module is useful.

Other approaches take advantage of different features of different domains. For instance, the "factored MDP" approach (used, for instance, by Boutilier et al., 2000, or Guestrin, Koller, Parr, & Venkataraman, 2003) is suitable for domains where parts of the state and action spaces can be grouped together so that within each group those actions or action dimensions affect the corresponding states or state dimensions but interaction between the groups is weak. St-Aubin, Hoey, and Boutilier (2000) iterate a symbolic representation in the form of algebraic decision diagrams to produce approximate solutions, while Sanner and Boutilier (2009) iterate a symbolic representation of a whole class of problems in a domain, using symbolic dynamic programming, first-order algebraic decision diagrams and linear value approximation, to pre-compute a generic solution which can then be used to quickly solve specific problems of that class. While we focus only on the state space, others approximate the action space, typically grouping actions (possibly hierarchically) into "macro actions", after Korf (1985). For instance Hauskrecht, Meuleau, Kaelbling, Dean, and Boutilier (1998) or Botea, Enzenberger, Muller, and Schaeffer (2005) take this approach, while Parr (1998) uses finite state automata for the macro actions and Srivastava, Immerman, and Zilberstein (2009) take it further by using algorithm-like plans with branches and loops. Goldman, Musliner, Boddy, Durfee, and Wu (2007) reduce the state space while generating the (limited-horizon, undiscounted) MDP from a different, non-MDP representation by only including reachable states, pruning those which can be detected as being clearly and immediately poor, or inferior or equivalent to already-generated states. Naturally, many of these approaches can be combined. For instance, Gardiol and Kaelbling (2004, 2008) combine state space abstraction with the envelope work of Dean et al. (1995), while Steinkraus (2005) uses a modular planner with a view of combining as many approaches as may be appropriate for a given problem. For more details and further approaches and variants we refer the reader to a recent survey of the field by Daoui, Abbad, and Tkiouat (2010).

### 2.6 Dynamic Approximate Planning

The top-level algorithm is shown as Algorithm 1. After some initialisation, consisting of selecting the initial abstraction and setting the policy, value and proximity to $a_0$, 0 and proportionally to the size of each worldview state, respectively,[5] the planner enters an infinite loop in which it stochastically alternates among five possible calculations, each of which is described in the following sections. Here and elsewhere in the algorithm, we use stochastic choice as a default in the absence of a more directed method.

The agent is assumed to have processing power available while it is acting, so that it can continually improve its policy, modify the approximation and updates the focus of its planning based on the current state. This means that the agent does not need to plan so well for unlikely possibilities, and can therefore expend more of its planning effort on the most likely paths and on the closer future, expecting that when and if it reaches other parts of the state space, it can improve the approximation as appropriate.

---

5. Initialising the approximate policy to action $a_0$ constitutes a domain-specific heuristic — namely, that there is a known default action $a_0$ which is reasonably safe in all states, such as a "do nothing" action.

---

**Algorithm 1** High-level algorithm for Approximate Planning with Dynamic Non-uniform Abstractions

  **do** select initial abstraction /* Algorithm 3 */
  **for all** worldview states $w$ **do**
    $\hat{\pi}(w) \leftarrow a_0$; $\hat{V}(w) \leftarrow 0$; $\wp\mathcal{P}(w) \leftarrow \frac{|w|}{|\mathcal{S}|}$
  **do** policy and value calculation /* Algorithm 2 */
  **loop**
    **choose** stochastically **do**
      **do** policy and value calculation /* Algorithm 2 */
    **or**
      **do** policy-based refinement /* Algorithm 4 */
    **or**
      **do** proximity calculation /* Algorithm 5 */
    **or**
      **do** proximity-based refinement /* Algorithm 6 */
    **or**
      **do** proximity-based coarsening /* Algorithm 7 */
    input latest current state; output the policy

---

Actual execution of the policy is assumed to be in a separate thread ('executive'), so that the planner does not have to concern itself with the timeliness requirements of the domain: whenever an action needs to be taken, the executive simply uses the policy that it most recently received from the planner.

Dean et al. (1995) call this *recurrent deliberation*, and use it with their locality-based approximation. A similar architecture is used by the CIRCA system (Musliner, Durfee, & Shin, 1995; Goldman, Musliner, Krebsbach, & Boddy, 1997) to guarantee hard deadlines. In CIRCA terminology, the planner is the AIS (AI subsystem), and the executive is the RTS (real-time subsystem).

An alternative to recurrent deliberation is *pre-cursor deliberation*, where the agent first plans, and only when it has finished planning does it begin to act, making no further adjustments to its plan or policy. Effectively, for the planner, the 'current state' is constant and equal to the initial state throughout planning. In this work the pre-cursor mode is used for some of the measurements, as it involves fewer potentially confounding variables.

Conceptually, our approach can be divided into two broad parts: the open-ended problem of selecting a good abstraction and the relatively closed problem of planning within that abstraction. Since the latter part is more closed, we deal with it first, in the next section, covering Algorithm 2. We then explore the more open-ended part in Sections 5–7, covering Algorithms 3–7.

## 3. Solving Non-uniformly Abstracted MDPs

Given a non-uniform abstraction, the simplest way to use it for planning is to take one of the standard MDP algorithms, such as the modified policy iteration of Puterman and Shin (1978), and adapt it to the non-uniform abstraction minimally. The formulae translate directly in the obvious fashion. $\hat{\pi}$ becomes a function of worldview states instead of concrete states, and so on, as shown in Algorithm 2 (using the simple variant for the **update policy for** $w$ procedure). Probabilities of transition from one worldview state to another are approximated using a uniform distribution over the concrete states (or possibly some other distribution, if more information is available).

---

**Algorithm 2** policy and value calculation

**repeat** $n$ times
  **for all** worldview states $w$ **do**
    **do** update value for $w$
**for all** worldview states $w$ **do**
  **do** update policy for $w$
  **do** update value for $w$

**procedure** update value for $w$
  **if** $\Pr_T(w, \hat{\pi}(w), w) = 1$ **then**
    /* optimisation — $\hat{V}(w)$ can be calculated directly in this case */
    $\hat{V}(w) \leftarrow \frac{R(w)}{1-\gamma}$
  **else**
    $\hat{V}(w) \leftarrow R(w) + \gamma \sum_{w'} \Pr_T(w, \hat{\pi}(w), w') \hat{V}(w')$

**procedure** update policy for $w$ **variant** simple
  $\hat{\pi}(w) \leftarrow \min \arg\max_a \sum_{w'} \Pr_T(w, a, w') \hat{V}(w')$

**procedure** update policy for $w$ **variant** with Locally Uniform Abstraction
  /* see Section 3.1 for discussion of Locally Uniform Abstraction */
  `absdims` $\leftarrow \{d : \exists w' \exists a \,.\, \Pr_T(w, a, w') > 0 \wedge w'$ is abstract in $d\}$
  `LUA` $\leftarrow \lambda w' \,.\, \bar{w}' : \begin{cases} \bar{w}' \text{ is abstract in } d & \forall d \in \texttt{absdims} \\ \text{dimension } d \text{ of } \bar{w}' = \text{dimension } d \text{ of } w' & \forall d \notin \texttt{absdims} \end{cases}$
  $\bar{V} \leftarrow \lambda \bar{w}' \,.\, \sum_{w'' \in \mathcal{W}} \frac{|w'' \cap \bar{w}'|}{|\bar{w}'|} \hat{V}(w'')$
  $\hat{\pi}(w) \leftarrow \min \arg\max_a \sum_{w'} \Pr_T(w, a, w') \bar{V}(\texttt{LUA}(w'))$

---

Note that in Algorithm 2, $A$ is considered an ordered set with $a_0$ as its smallest element and the minimum is used when the arg max gives more than one possibility. This has two aspects: (a) as a domain-specific heuristic, for instance, breaking ties in favour of the default action when possible, and (b) to avoid policy-based refinement (see Section 5) based on actions that have equal value. Secondly, for efficiency, $\sum_{w'}$ can be calculated only over states $w'$ with $\Pr_T(w, a, w') > 0$, since other states will make no contribution to the sum. Finally, the number $n$ is a tuning parameter which is not particularly critical (we use $n = 10$).

Of course, replacing the state space $\mathcal{S}$ by a worldview $\mathcal{W}$ in this way does not, in general, preserve the Markov property, since the actual dynamics may depend on aspects of the state space that are abstracted in the worldview. In the simple variant we ignore this and assume the Markov property anyway, on the grounds that this is, after all, an approximation. Unfortunately, the resulting performance can have unacceptably large error, including the outright non-attainment of goals.

For instance, in the `3Doors` problem, such a situation will occur at each of the three doors whenever they are all abstract at $s^0$ and concrete near the door in question. The doors are relatively difficult to open, with only a 10% probability of success per try. On the other hand, when moving from an area where they are abstract to an area where they are concrete, the assumed probability that the door is already open is 50%. When the calculations are performed, it turns out to be preferable to plan a loop, repeatedly trying for the illusory 50% chance of success rather than attempting to open the door at only 10% chance of success. The agent will never reach the goal. Worse still, in some ways, it will estimate that the quality of the solution is quite good, $\hat{V}_{\hat{\pi}}(s^0) \approx -19.0$, which is in fact better even than the optimal solution's $V^*(s^0) \approx -27.5$, while the true quality of the solution is very poor, $V^*_{\hat{\pi}}(s^0) = -100\,000$, corresponding to never reaching the goal (but not incurring damage, either; figures are for discounting factor $\gamma = 0.999\,99$).

Regions that take into account a particularly bad piece of information may seem unattractive, as described above, and vice versa. We call this problem the *Ostrich effect*, as the agent is refusing to accept an unpleasant fact, like the mythical ostrich that buries its head in the sand. Its solution, *Locally Uniform Abstraction*, is described in the next section.

If the abstracted approximation is simply treated as a MDP in which the agent does not know which state it will reach (near closed door or near open door), it will not correspond to the underlying process, which might reach a particular state deterministically (as it does here). The problem is especially obvious in this example, when the planner plans a loop. This is reminiscent of a problem noted by Cassandra, Kaelbling, and Kurien (1996), where a plan derived from a POMDP failed — the actual robot got into a loop in a particular situation when a sensor was completely reliable contrary to the model.

### 3.1 Locally Uniform Abstraction

The ostrich effect occurs when states of different abstraction are considered, for instance one where a door is abstract and one where the same door is concrete and closed. The solution is to make the abstraction locally uniform, and therefore locally Markovian for the duration of the policy generation iterative step. By making the abstraction locally, temporarily uniform, the iterative step of the policy generation algorithm never has to work across the edge of an abstract region, and, since the same information is available in all the states being considered at each point, there is no impetus for any of them to be favoured or avoided on that basis (for instance, avoiding a state in which a door is concrete and closed in favour of one where the door is abstract). The action chosen will be chosen based on the information only and not on its presence or absence.

This is a modification to the **update policy for** $w$ procedure of Algorithm 2: as the states are considered one by one, the region around each state is accessed through a function that returns a locally uniform version. States that are more concrete than the state being considered will be averaged so as to ignore the distinctions. As different states are considered, sometimes the states will be taken for themselves, sometimes their estimated values $\hat{V}$ will be averaged with adjacent states. This means that some of the dimensions will only partially be considered at those states — in most cases, this will mean that the more concrete region must extend one step beyond the region in which the dimension is immediately relevant. For a dimension to be fully considered at a state, the possible outcomes of all actions at that state must also be concrete in that dimension.

The modified procedure proceeds as follows: first the dimensions that are abstract in any possible outcome of the state being updated $w$ are collected in the variable `absdims`. Then the function `LUA` is constructed which takes worldview states $w'$ and returns potential worldview states $\bar{w}'$ which are like $w'$ but abstract in all the dimensions in `absdims`. As this is the core of the modification, it is named `LUA` for 'Locally Uniform Abstraction'. Since the potential states returned by `LUA` are not, in general, members of $\mathcal{W}$, and therefore do not necessarily have a value stored in $\hat{V}$, a further function $\bar{V}$ is constructed which calculates weighted averages of the value function $\hat{V}$ over potential states. As with the other sum, $\sum_{w''}$ can be calculated only over states $w''$ with $w'' \cap \bar{w}' \neq \emptyset$ for efficiency. Finally, the update step is carried out using the two functions `LUA` and $\bar{V}$.

Unfortunately, once the modification is applied, the algorithm may or may not converge depending on the worldview. Failure to converge occurs when the concrete region is too small — in some cases, the algorithm will cycle between two policies (or conceivably more) instead of converging. One must be careful, therefore, with the worldview, to avoid these situations, or else to detect them and modify the worldview accordingly. The policy-based worldview refinement algorithm described in Section 5 below ensures convergence in practice.

## 4. Initial Abstraction

At the beginning of planning, the planner must select an initial abstraction. Since the worldview is never completely discarded by the planner, an infelicity at this stage may impair the entire planning process, as the worldview-improvement algorithms can only make up for some amount the weakness here.

There are different ways to select the initial abstraction. We propose one heuristic method for selecting the initial worldview based on the problem description, with some variants. Consider for example that each door in the `3Doors` problem is associated with two locations, that is, those immediately on either side. It makes sense, then, to consider the status of the door in those two locations. This association can be read off the problem specification. Intuitively, the structure of the solution is likely to resemble the structure of the problem. This incorporates the structure of the transition function into the initial worldview. The reward function is also incorporated, reflecting the assumption that the dimensions on which the reward is based will be important.

We use a two-step method to derive the initial worldview, as shown in Algorithm 3. Firstly, the reward function is specified based on particular dimensions. We make those dimensions concrete throughout the worldview, and leave all other dimensions abstract. In the `3Doors` problem, these are the `x` and `y` and `dmg` dimensions, so after this step there are $10 \times 10 \times 2 = 200$ states in the worldview.

Secondly, the transition function is specified by decision trees, one per action. We use these to find the *nexuses* between the dimensions, that is, linking points, those points at which the dimensions interact. Each nexus corresponds to one path from the root of the tree to a leaf. For example, in the `3Doors` problem, the decision tree for the "open" action contains a leaf whose ancestors are $x$, $y$, `d1` and a stochastic node, with the choices leading

**Algorithm 3** select initial abstraction

/* set the worldview completely abstract */
$\mathcal{W} \leftarrow \{\mathcal{S}\}$
/* reward step */
**if** reward step enabled **then**
    **for all** dimensions $d$ mentioned in the reward tree **do**
        refine the whole worldview in dimension $d$
/* nexus step */
**if** nexus step enabled **then**
    **for all** leaf nodes in all action trees **do**
        **for all** worldview states $w$ matching the pre-state **do**
            refine $w$ in the dimensions mentioned in the pre-state

to that leaf being labelled respectively 4, 2, closed and 10%. This corresponds to a nexus at $s_x = 4$, $s_y = 2$ and $s_{\texttt{d1}} = $ closed (the stochastic node is ignored in determining the nexus). In total, there are four nexuses on each side of each door, in the two locations immediately adjacent, as shown in Figure 2(a), connecting the relevant door dimension to the x and y coordinates. The initial worldview is shown in Figure 2(b), with x, y and dmg concrete everywhere and the doors abstract except that each is concrete in the one location directly on each side of the door, corresponding to the location of the nexuses on Figure 2(a). After both steps, $|\mathcal{W}| = 212$, compared to $|\mathcal{S}| = 1\,600$ specific states.

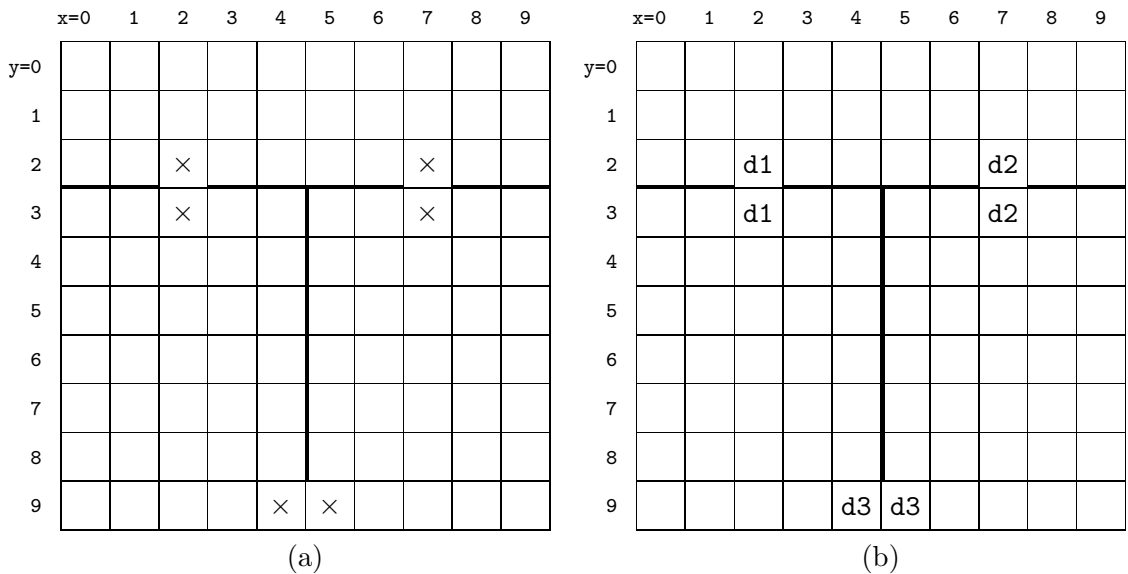

Figure 2: Nexus step of the initial abstraction, showing (a) the location of the nexuses in the 3Doors problem (there are four nexuses at each ×) and (b) the locations in which the door dimensions will be concrete in the initial worldview.

For the 3Keys problem, the location of the nexuses is the same as in Figure 2(a), except there are more nexuses in each location and some of them also involve the corresponding key dimensions. Thus, in the initial worldview, the locations shown in Figure 2(b) will be concrete not only in corresponding door dimension, but also, when they are closed, in the corresponding key dimension. In the states in which the doors are open, the key dimension remains abstract. The initial worldview size for 3Keys is $|\mathcal{W}| = 224$.

Due to the locally-uniform abstraction, these concrete door dimensions will be taken into account only to a very minimal degree. If the worldview were to be used without further refinement, it is to be expected that the resulting policies would be very poor. The results[6] bear out this expectation. The worldview initialization methods therefore are not intended to be used on their own, but rather as the basis for further refinement. Thus, the real test of the methods is how well they will work when coupled with the worldview modification methods, described below.

## 5. Policy-Based Refinement

This section presents the first of the worldview modification methods, policy-based refinement. This method modifies the worldview based directly on the current approximate policy $\hat{\pi}$. In particular, it refines states based on differences between the actions planned at adjacent, differently-abstract states. Where such differences indicate that a dimension may be important, any adjacent states that are abstract in that dimension are refined (i.e. that dimension is made concrete at those states).

The method was previously introduced by Baum and Nicholson (1998), who showed, using a small navigation domain example (the 3Doors problem of this paper), that this refinement method resulted in a good policy, though not optimal. Here we present quantitative results and consider more complex domains.

### 5.1 Motivation

The motivation for this method is twofold. Firstly, as already indicated, the method detects areas in which a particular dimension is important, because it affects the action planned, and ensures that it is concrete at adjacent states. Thus regions where a dimension is taken into account will expand for as long as the dimension matters, and then stop. Secondly, the method fulfils the requirements for choosing a worldview so as to avoid non-convergence in the policy calculation, as mentioned in Section 3.1 above.

Dimensions are important where they affect the policy, since the policy is the planner's output. They are less important in parts of the state space where they do not affect the policy. Thus, which dimensions need to be concrete and which can remain abstract can be gleaned for each part of the state space by comparing the optimal actions for the various states. Where the optimal actions are equal, states can be abstract, and where they differ, states should be concrete. However, we do not have the optimal policy $\pi^*$. With an approximate policy $\hat{\pi}$ on a worldview, it is more difficult. However, the planner can compare policies in areas where a dimension is concrete, and if it is found to be important there, expand the area in which it is concrete. As policy-based refinement and policy calculation alternate, refinement will continue until the area where the dimension is concrete covers the whole region in which it is important.

---

6. Omitted here as they uninteresting, but presented by Baum (2006).

Section 3.1 above noted that the planning algorithm requires a worldview chosen with care. The algorithm described in this section detects situations that are potentially problematic under locally-uniform abstraction and modifies the worldview to preclude them. Intuitively, the incorrect behaviour occurs where an 'edge' of a concrete region intersects with a place where there are two fairly-similarly valued courses of action, corresponding to two different paths to the goal.

## 5.2 Method

The method uses the transition function as the definition of adjacent states, so that worldview states $w$ and $w'$ are considered adjacent if $\exists a \,.\, \Pr_T(w, a, w') > 0$. This definition is not symmetrical in general, since the transition function is not, but that is not a problem for this method, as can be seen below. The algorithm is shown as Algorithm 4.

---

**Algorithm 4** policy-based refinement

$\quad$ candidates $\leftarrow \emptyset$
$\quad$ **for all** worldview states $w$ **do**
$\quad\quad$ **for all** actions $a$ **do**
$\quad\quad\quad$ **for all** $w' : Pr(w, a, w') > 0$ **do**
$\quad\quad\quad\quad$ **for all** dimensions $d : w$ is abstract in $d$ **and** $w'$ is concrete in $d$ **do**
$\quad\quad\quad\quad\quad$ construct $\bar{w}' : \begin{cases} \bar{w}' \text{ is abstract in } d \\ \text{dimension } d' \text{ of } \bar{w}' = \text{dimension } d' \text{ of } w' \quad \forall d' \neq d \end{cases}$
$\quad\quad\quad\quad\quad$ **if** $\exists w^a, w^b \,.\, \hat{\pi}(w^a) \neq \hat{\pi}(w^b)$ **and** $w^a \cap \bar{w}' \neq \emptyset$ **and** $w^b \cap \bar{w}' \neq \emptyset$ **then**
$\quad\quad\quad\quad\quad\quad$ /* policy is not the same throughout $\bar{w}'$ */
$\quad\quad\quad\quad\quad\quad$ candidates $\leftarrow$ candidates $\cup \{(w, d)\}$
$\quad$ **for all** $(w, d) \in$ candidates **do**
$\quad\quad$ **if** $w \in \mathcal{W}$ **then**
$\quad\quad\quad$ /* replace $w$ with a group of states concrete in $d$ */
$\quad\quad\quad$ **for all** $w^{new} : \begin{cases} w^{new} \text{ is concrete in } d \\ \text{dimension } d' \text{ of } w^{new} = \text{dimension } d' \text{ of } w \quad \forall d' \neq d \end{cases}$ **do**
$\quad\quad\quad\quad$ $\mathcal{W} \leftarrow \mathcal{W} \cup \{w^{new}\}$
$\quad\quad\quad\quad$ $\hat{\pi}(w^{new}) \leftarrow \hat{\pi}(w); \hat{V}(w^{new}) \leftarrow \hat{V}(w); \mathcal{P}_{\hat{\pi}}(w^{new}) \leftarrow \frac{|w^{new}|}{|w|} \mathcal{P}_{\hat{\pi}}(w)$
$\quad\quad\quad$ $\mathcal{W} \leftarrow \mathcal{W} \setminus \{w\}$ /* discarding also the stored $\hat{\pi}(w)$, $\hat{V}(w)$ and $\mathcal{P}_{\hat{\pi}}(w)$ */

---

**Example** In the `3Doors` problem, for instance, applying this method during planning increases the number of worldview states from the initial 212 to 220–231, depending on the stochastic choices (recall that $|\mathcal{S}| = 1\,600$ for comparison). It produces concrete regions which are nice and tight around the doors, as shown in Figure 3, while allowing the algorithm to converge to a reasonable solution. The solution is in fact optimal for the given initial state $s^0$, though that is simply a coincidence, since the $s^0$ is not taken into account by the algorithm and some other states have somewhat suboptimal actions (the agent would reach the goal from these states, but not by the shortest route).

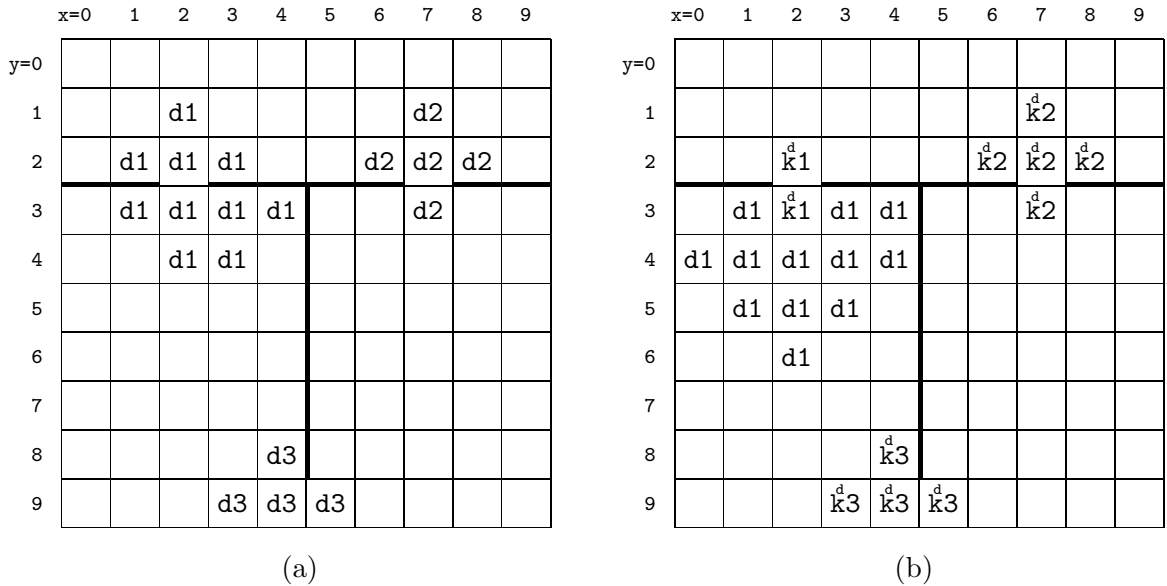

Figure 3: Example of non-uniform abstraction for the (a) 3Doors and (b) 3Keys problems with policy-based refinement. The x, y and dmg dimensions are concrete everywhere; d1, d2 and d3 indicate where the corresponding door is concrete; while $\overset{d}{k}1$, $\overset{d}{k}2$ and $\overset{d}{k}3$ indicate that the corresponding door is concrete and the corresponding key is also concrete if the door is closed.

The worldview obtained by this method is often quite compact. For instance, rather than refining a simple $2 \times 3$ rectangular region on each side of a door in 3Doors, as a human might, this algorithm makes only 4 locations concrete on the approach side of each door, which is enough to obtain a good solution. This can be seen on the north sides of doors d1 and d2, as well as the west side of door d3 (3 concrete locations, due to the edge). On the departure side of doors d2 and d3, it is even better — it makes no refinement at all: south of door d2 and east of door d3, the action is to move toward the goal, regardless of the status of the door — the actions are equal, so no refinement takes place.

The south side of door d1 seems rather less compact. The concrete area is in fact not very big — 6 locations for 3Doors — but it seems excessive compared with the compact concrete areas elsewhere. This can occur when there is a nexus close to a region where the best action to take genuinely depends on the status of the dimension in the nexus, but the difference is small. If somehow the agent found itself at $\langle 4, 3 \rangle$ — and policy-based refinement is independent of $s^{\text{cur}}$ — the optimal path genuinely would depend on whether door d1 is open, the other path being slightly suboptimal in each case. While in theory such a region could have arbitrarily large extent, it seems to be a relatively minor effect in practice. Here, for instance, it adds a couple of states, which is about 1% of $|\mathcal{W}|$, and it was not found to be a real problem in any of the domains (or in the domains used in Baum, 2006).

### 5.3 Limitations

Policy-based refinement can only deal with cases where a single dimension makes a difference. When two dimensions are needed in combination, it will often miss them. For instance, in the 3Keys problem each key is quite distant from the corresponding door and policy-based refinement will therefore never find the relationship between the two. At the key, there appears to be no reason to pick it up, while at the door there appears to be no means of unlocking it.

Obviously, this can be fixed *ad hoc* by rewarding picking up keys for its own sake. Indeed, some domain formulations in the literature do exactly that, rewarding the agent for partial achievement of the goal. However, that is not a clean solution. In effect, such domain specifications 'cheat' by providing such hints.

Another problem is that policy-based refinement does not provide for coarsening the worldview, or for modifying it in other ways, for instance as execution progresses and the planner needs to update the plan. Indeed, policy-based refinement ignores the initial state $s^0$ altogether, or the current state $s^{\mathrm{cur}}$ in recurrent planning. Thus it produces the same solution regardless of which part of the problem the agent has actually been asked to solve. This is a waste of computation in solving those parts which the agent is unlikely to actually visit, and — perhaps more importantly — carries the penalty of the corresponding loss of quality in the relevant parts.

The following sections describe proximity-based worldview modification, which is needed to solve domains where combinations of dimensions are important and which also makes use of $s^0$ or $s^{\mathrm{cur}}$, as appropriate.

## 6. A Proximity Measure

In general, the worldview should be mostly concrete near the agent and its planned path to the goal, to allow detailed planning, but mostly abstract elsewhere, to conserve computational resources. In this section we describe a measure (originally in Baum & Nicholson, 1998) which realises this concept, *proximity* $\mathcal{P}$, which decreases both as a state is further in the future and as it is less probable.[7] This section extends the brief description of Baum and Nicholson (1998). In the following section we then present new worldview modification methods based directly on the measure.

### 6.1 Motivation

The proximity $\mathcal{P}$ is a realisation of the intuitive concept of states being near the agent and likely to be visited, as opposed to those distant from the agent and unlikely. It naturally takes into account the current state $s^{\mathrm{cur}}$ in recurrent planning, or the initial state $s^0$ in pre-cursor planning, unlike policy-based refinement which ignores them altogether. Thus a planner selecting worldviews based on this proximity measure will produce solutions tailored to the particular $s^{\mathrm{cur}}$ or $s^0$ and will ignore parts of the MDP that are irrelevant or near-irrelevant to performance from that state. Thus it saves computation that would otherwise be wasted in solving those parts which the agent is unlikely to actually visit, and — perhaps more importantly — carries the advantage of the corresponding gain of quality in the relevant parts. This allows the agent to deal with problems such as `3Keys` which are beyond the reach of policy-based refinement.

---

7. Baum and Nicholson (1998) used the word 'likelihood' for this measure. We now prefer 'proximity' to avoid confusion with the other meanings of the word 'likelihood'. Munos and Moore (1999) use the word 'influence' for a somewhat similar measure in continuous domains.

Implicitly, the agent plans that when and if it reaches those mostly-abstract parts of the state space, it will improve the approximation as appropriate. The planner thus continually improves the policy, modifies the approximation and updates the focus of its planning based on the current state $s^{\text{cur}}$. This means refining the regions in which the agent finds itself or which it is likely to visit, and coarsening away details from regions that it is no longer likely to visit or those which it has already traversed.

There are three aspects to proximity: temporal, spatial and probabilistic. Firstly, the temporal aspect indicates states that may be encountered in the near future, on an exponentially decaying scale. The second aspect is spatial — the nearness of states (in terms of the state space) to the agent and its planned path. The spatial aspect is somewhat indirect, because any spatial structure of the domain is represented only implicitly in the transition matrix, but the proximity measure will reflect it. These two aspects are combined in the proximity to give a single real number between 0 and 1 for each state, denoted $^\ell_\tau\mathcal{P}$ — $\mathcal{P}$ for proximity, $\ell$ for the spatial aspect and $\tau$ for the temporal aspect. This number can be interpreted as a probability — namely the probability of encountering a state — and $^\ell_\tau\mathcal{P}$ can be interpreted as the probability distribution over states, giving the final, probabilistic aspect of proximity.

### 6.2 Calculation

The formula for the proximity $^\ell_\tau\mathcal{P}$ is similar to the formula for the value function. There are three differences. Firstly, instead of beginning from the reward function it is based on an 'is current state' function, *cur*. Secondly, the transition probabilities are time-reversed (that is, the matrix is transposed). This is because the value calculation is based on the reward function, which occurs in the future (after taking actions), while the 'is current state' function is based on the present, before taking actions. Since the order of taking actions and the function upon which the formula is based is reversed in time, a similar reversal must be applied to the transition probabilities. Thirdly, an estimated future policy $\widehat{\hat{\pi}}$ is used instead of $\hat{\pi}$. In this estimate, $\widehat{\hat{\pi}}$ is a stochastic policy defined by making $\widehat{\hat{\pi}}(s)$ a distribution over actions which assigns some constant probability to the current $\hat{\pi}(s)$ and distributes the remaining probability mass among the other actions equally. This distributed probability mass corresponds to the probability that the policy will change sometime in the future, or, alternately, the probability that the currently-selected action is not yet correct. The formula is therefore:

$$^\ell_\tau\mathcal{P}(s) \leftarrow \text{cur}(s) + \gamma_\mathcal{P} \sum_{s'} \Pr_T(s', \widehat{\hat{\pi}}(s'), s) ^\ell_\tau\mathcal{P}(s') \tag{1}$$

where

$\gamma_\mathcal{P}$ is the proximity discounting factor $(0 \leq \gamma_\mathcal{P} < 1)$

and

$$\text{cur}(s) = \begin{cases} 1 - \gamma_\mathcal{P} & \text{if } s^{\text{cur}} = s \\ 0 & otherwise \end{cases} \tag{2}$$

The constant $1-\gamma_\mathcal{P}$ was chosen for the current-state function so that $\sum_s {}^\ell_\tau\mathcal{P}(s)$ converges to 1, in other words so that ${}^\ell_\tau\mathcal{P}$ is a probability distribution. If checked in the near future, the agent has a probability of ${}^\ell_\tau\mathcal{P}(s)$ of being in state $s$, assuming it will follow the policy $\hat{\pi}$ and 'near future' is defined so that the probability of checking at time $t$ is proportional to $\gamma_\mathcal{P}^t$ (that is, with $\gamma_\mathcal{P}$ interpreted as a stopping probability). As with the value calculation, one can instead solve the set of linear equations

$${}^\ell_\tau\mathcal{P}(s) = \mathrm{cur}(s) + \gamma_\mathcal{P} \sum_{s'} \Pr_T(s', \widehat{\hat{\pi}}(s'), s) {}^\ell_\tau\mathcal{P}(s') \tag{3}$$

or, in matrix notation,

$$(I - \gamma_\mathcal{P} T_{\widehat{\hat{\pi}}}^T) {}^\ell_\tau\mathcal{P} = \mathrm{cur} \tag{4}$$

where $T_{\widehat{\hat{\pi}}}$ is the transition matrix induced by the stochastic policy $\widehat{\hat{\pi}}$ and $I$ is the identity matrix. The implementation uses this matrix form, as shown in Algorithm 5. The proximity measure needs little adjustment to work with the non-uniformly abstract worldview: $s$ is simply replaced by $w$ in (1) and (2), with $s^{\mathrm{cur}} = s$ becoming $s^{\mathrm{cur}} \in w$.

---

**Algorithm 5** proximity calculation

    solve this matrix equation for ${}^\ell_\tau\mathcal{P}$ as a linear system:

$$(I - \gamma_\mathcal{P} T_{\widehat{\hat{\pi}}}^T) {}^\ell_\tau\mathcal{P} = \mathrm{cur}$$

---

The measure has two tuning parameters, the replanning probability and the discounting factor $\gamma_\mathcal{P}$. The replanning probability controls the spatial aspect: it trades off focus on the most likely path and planning for less likely eventualities nearby. Similarly, $\gamma_\mathcal{P}$ controls the temporal aspect: the smaller $\gamma_\mathcal{P}$ is, the more 'short sighted' and greedy the planning will be. Conversely, if $\gamma_\mathcal{P}$ is close to 1, the planner will spend time planning for the future that might have been better spent planning for the here-and-now. This should be set depending on the reward discounting factor $\gamma$, and on the mode of the planner. Here we use $\gamma_\mathcal{P} = 0.95$, replanning probability 10%.

**Example** Proximities for the 3Doors problem are shown in Figure 4 for the initial situation (agent at $\langle 0, 0 \rangle$, all doors closed) and a possible situation later in the execution (agent at $\langle 4, 2 \rangle$, all doors closed). Larger symbols correspond to higher proximity. One can immediately see the agent's planned path to the goal, as the large symbols correspond to states the agent expects to visit. Conversely small proximities show locations that are not on the agent's planned path to the goal. For example, the agent does not expect to visit any of the states in the south-western 'room', especially once it has already passed by door 1. Similarly, the proximities around the initial state are much lower when the agent is at $\langle 4, 2 \rangle$, as it does not expect to need to return.

### 6.3 Discussion

One interesting feature of the resulting numbers is that they emphasise absorbing and near-absorbing states somewhat more than might be intuitively expected. However, considering that absorbing states are in general important, this is a good feature, especially since normally the planner will try to minimise the probability of entering an absorbing state (unless it is the goal). This feature should help ensure that absorbing states are kept in mind as long as there is any chance of falling into them. Dean et al. (1995), for instance, note that in their algorithm such undesirable absorbing states along the path to the goal tend to come up as candidates for removal from consideration (due to the low probability of reaching them with the current policy), and have to make special accommodation for them so they are not removed from consideration. With the proximity measure emphasising these states, such special handling is not necessary.

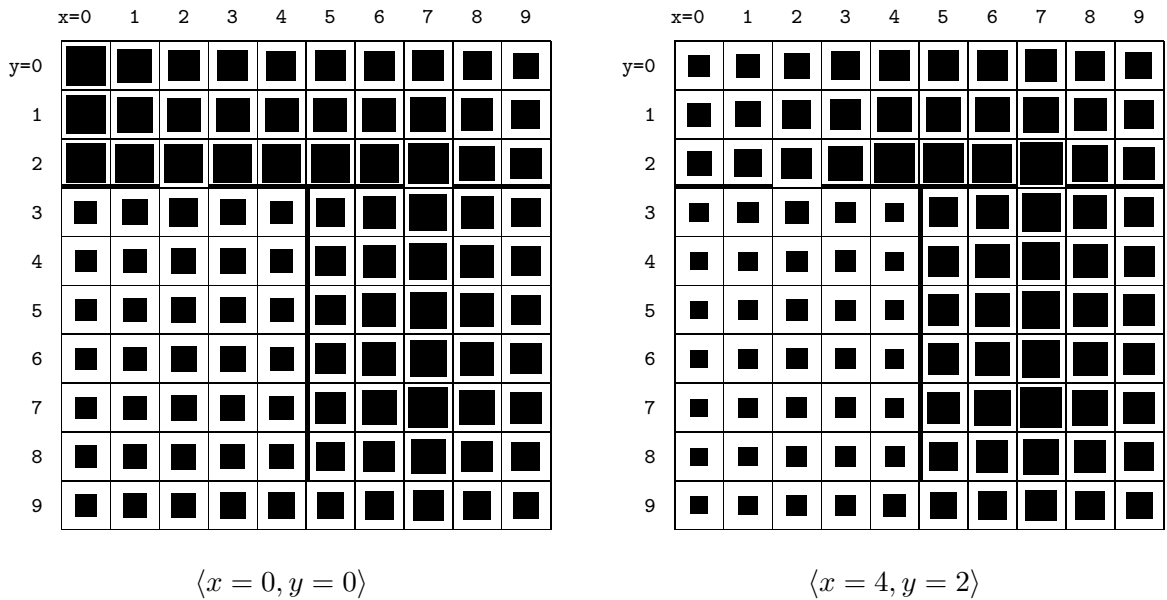

Figure 4: Proximities in the 3Doors problem for $s^0$ and a possible later $s_{cur}$; symbol size logarithmic, proximities range from $2^{-37}$ to $2^{-1.4}$; $\gamma_\mathcal{P} = 0.95$, replanning probability 10%.

In contrast with this approach, Kirman (1994) uses the probabilities after $E_s$ steps, where $E_s$ is (an estimate of) the number of steps the agent will take before switching from the previous policy to the policy currently being calculated. This assumes that $E_s$ can be estimated well, that the current policy is the policy in the executive, and that the one-planning-cycle probability is an appropriate measure. In fact one would prefer at least a two-planning-cycle look-ahead, so that the agent not only begins within the area of focus of the new policy, but also remains there throughout the validity of that policy, and probably longer, since the planner's foresight should extend beyond its next thinking cycle. More philosophically, this reliance on the planning cycle length is not very desirable, as it is an artefact of the planner rather than intrinsic in the domain.

A somewhat related approach is *prioritised sweeping* (see for instance Barto, Bradtke, & Singh, 1995). Like the present approach, it defines a measure of states which are in some way 'interesting'. Unlike this approach, it then applies that measure to determine the order in which the formulae of the $\hat{\pi}$ and $\hat{V}$ calculation are applied, so that they are applied preferentially to the interesting states and less frequently to the uninteresting or unimportant states. It is well-known that the order of calculation in the MDP planning algorithms can be varied greatly without forfeiting convergence to the optimal policy, and prioritised sweeping takes advantage of this. Often it is done on a measure such as change in $\hat{V}$ in previous calculations, but some approaches use look-ahead from the current state, which is in some ways a very simple version of proximity (in fact, it corresponds to a threshold on $\mathcal{P}$ with replanning probability set to 1). The proximity measure $\mathcal{P}$ might well be a good candidate for this approach: apply $\hat{\pi}$ and $\hat{V}$ calculation to states chosen directly according to $\mathcal{P}$ as a distribution.[8]

Munos and Moore (1999) use an 'influence' measure on their deterministic continuous domains, which is very similar to $\mathcal{P}$. In fact, the main difference is that their measure does not have the two parameters — it re-uses the same $\gamma$ and has no replanning probability (effectively it is zero). This means that it cannot take into account replanning, neither in the difference in the horizon that it entails, nor in the possibility that the policy may change before it is acted upon. Absorbing states, for instance, would not be emphasised as they are with proximities.

## 7. Proximity-Based Dynamic Abstraction

The proximity measure described in the previous section is used to focus the planner's attention to areas likely to be useful in the near future. Firstly, that means that the worldview should be made to match the proximities, by refining and coarsening as appropriate. Secondly, since the proximity measure takes into account the current state, this method will automatically update the worldview as the agent's circumstances change in the recurrent mode, that is, when planning and execution are concurrent.

### 7.1 Refinement

High proximity indicates states which the agent is likely to visit in the near future. The planner should therefore plan for such states carefully. If they are abstract, this is reason to refine them so as to allow more detailed planning. Such states with high proximity are therefore considered candidates for refinement.

High proximity is defined by a simple threshold, as shown in Algorithm 6.

When refinement occurs, an anomaly sometimes appears. Like the anomaly which led to the policy-based refinement method, it arises from different levels of abstraction, but here, it is not an adjacent more abstract state that causes the problem, but rather a recently-refined one. When a state is refined, the values $\hat{V}$ of the new states are initially estimated from the state's previous value $\hat{V}$. However, typically, this means that some of them will be overestimated and others underestimated. When the policy is being re-calculated, the state with the overestimated value will be attractive.

Since the problem directly follows from the moment of refinement, it is self-correcting. After a few iterations, the planner converges to the correct policy and values. However,

---

8. If retaining the theoretical guarantee of convergence to $\pi^*$ is desired, care would have to be taken since $\mathcal{P}$ is zero for states which are not reachable from the current state. In practice, of course, optimality or otherwise as to unreachable states is immaterial.

---

**Algorithm 6** proximity-based refinement
  stochastically choose a dimension $d$
  **for all** worldview states $w$ **do**
    **if** $\wp\mathcal{P}(w) >$ threshold **and** $w$ is abstract in $d$ **then**
      /* replace $w$ with a group of states concrete in $d$ */
      **for all** $w^{new} : \begin{cases} w^{new} \text{ is concrete in } d \\ \text{dimension } d' \text{ of } w^{new} = \text{dimension } d' \text{ of } w \quad \forall d' \neq d \end{cases}$ **do**
        $\mathcal{W} \leftarrow \mathcal{W} \cup \{w^{new}\}$
        $\hat{\pi}(w^{new}) \leftarrow \hat{\pi}(w);\ \hat{V}(w^{new}) \leftarrow \hat{V}(w);\ \wp\mathcal{P}(w^{new}) \leftarrow \frac{|w^{new}|}{|w|}\wp\mathcal{P}(w)$
      $\mathcal{W} \leftarrow \mathcal{W} \setminus \{w\}$ /* discarding also the stored $\hat{\pi}(w), \hat{V}(w)$ and $\wp\mathcal{P}(w)$ */

---

while it is doing so, transient anomalies appear in the policy, and in the worst case, the planner may replan for some other path, then refine more states and re-trigger the same anomaly. Rather large parts of the state space can be spuriously refined in this way.

This occurs because of the combined $\hat{\pi}$ and $\hat{V}$ calculation phase, which may update $\hat{\pi}$ before $\hat{V}$ has had a chance to converge. The solution is to create a variant phase, $\hat{V}$ calculation only, which replaces the $\hat{\pi}$ and $\hat{V}$ calculation phase until the values stabilise. We do this for two iterations, which appears to be sufficient. An alternative solution would have been to copy the difference between values at adjacent, more concrete states where possible, thus obtaining better estimated values at the newly-refined states. However, since the simpler solution of $\hat{V}$-only calculation works satisfactorily, this more complex possibility has not been further explored.

### 7.2 Coarsening

Low proximity indicates states which the agent is unlikely to visit in the near future. The planner therefore need not plan for such states carefully. Usually, they will already be abstract, never having been refined in the first place. However, if they are concrete — if they have been previously refined — this is reason to coarsen them so as to free up memory and CPU time for more detailed planning elsewhere. Such states with low proximity are therefore considered candidates for coarsening.

Proximity-based coarsening is useful primarily in an on-line planning scenario with recurrent planning. As the agent itself moves through the state space and the current state $s^{\text{cur}}$ changes, so do the states that are likely to be visited in the near future. This is especially useful if the agent finds itself in some unexpected part of the state space, for instance due to low-probability outcomes, or if the agent has planned a path leading only part way to the goal (perhaps up to a partial reward). In any case, however, the parts of the state space already traversed can be coarsened in favour of refinement in front of the agent.[9]

One might also imagine that as planning progresses, the planner may wish to concentrate on different parts of the state space and that coarsening might be useful to cull abandoned explorations and switch focus. However, we have not observed this with any of our domains and found that in pre-cursor mode, coarsening generally worsens the quality of the policies with no positive contribution.

---

9. States already traversed cannot be discarded, even if the agent will never visit them again, since the worldview is a partition and since the agent does not necessarily know whether it will need to revisit (or end up revisiting) those states.

Coarsening proceeds in three steps, as shown in Algorithm 7. The first step is very similar to proximity-based refinement: each time the 'proximity-based coarsening' phase is invoked, the worldview is scanned for states with low proximity (below a threshold), which are put in a list of candidates. The second step is more tricky. Coarsening needs to join several states into one. However, the representation does not allow arbitrary partitions as worldviews and therefore does not allow the coarsening-together of an arbitrary set of worldview states. The planner must therefore find a group of states among these low-proximity candidates which can be coarsened into a valid worldview state. Such groups can be detected by the fact that they differ in one dimension only and have the same size as the dimension, therefore covering it completely. Finally, the groups that have been found are each replaced with a single more abstract state.

---

**Algorithm 7** proximity-based coarsening

/* collect candidates for coarsening */
candidates ← $\{w : \wp\mathcal{P} < \text{threshold}\}$

/* find groups of candidates that can be coarsened together */
/* partition candidates according to the pattern of abstract and concrete dimensions */
patterns ← candidates / $\{(w^a, w^b) : \forall d . w^a \text{ is concrete in } d \leftrightarrow w^b \text{ is concrete in } d\}$
groups ← $\emptyset$
**for all** p ∈ patterns **do**
  **for all** dimensions $d$ **do**
    **if** states in p are concrete in $d$ **then**
      /* partition p by all dimensions except $d$, giving potential groups */
      potgroups ← p / $\{(w^a, w^b) : \forall d' \neq d . \text{dimension } d' \text{ of } w^a = \text{dimension } d' \text{ of } w^b\}$
      /* add all potential groups that are the same size as dimension $d$ to groups */
      groups ← groups $\cup \{\text{g} \in \text{potgroups} : |\text{g}| = |\mathcal{S}_d|\}$

/* replace each group of states with a single, more abstract state */
**for all** g ∈ groups **do**
  **if** g $\subseteq \mathcal{W}$ **then**
    stochastically choose $w^a$ from g
    construct $w^{new}$ : $\begin{cases} w^{new} \text{ is abstract in } d \\ \text{dimension } d' \text{ of } w^{new} = \text{dimension } d' \text{ of } w^a \quad \forall d' \neq d \end{cases}$
    $\mathcal{W} \leftarrow \mathcal{W} \cup \{w^{new}\}$
    $\hat{\pi}(w^{new}) \leftarrow \hat{\pi}(w^a); \hat{V}(w^{new}) \leftarrow \frac{1}{|\text{g}|} \sum_{w \in \text{g}} V(w); \wp\mathcal{P}(w^{new}) \leftarrow \sum_{w \in \text{g}} \wp\mathcal{P}(w)$
    $\mathcal{W} \leftarrow \mathcal{W} \setminus \text{g}$ /* discarding also the stored $\hat{\pi}(w)$, $\hat{V}(w)$ and $\wp\mathcal{P}(w)$ for all $w \in g$ */

---

In some cases, it may be impossible to coarsen a section of the worldview despite low proximity, due to a situation somewhat akin to grid-lock. Probably the simplest example is the one in Figure 5, which shows a worldview of five states in a three-dimensional binary specific state space, with three of the states ignoring a different dimension each, while the remaining two take into account all three. In this situation, no group of states can be

|       |       |       |       |       |       |       |       |       |
|-------|-------|-------|-------|-------|-------|-------|-------|-------|
| $\mathcal{S}_1$ | 0 | 0 | 0 | 0 | 1 | 1 | 1 | 1 |
| $\mathcal{S}_2$ | 0 | 0 | 1 | 1 | 1 | 1 | 0 | 0 |
| $\mathcal{S}_3$ | 0 | 1 | 1 | 0 | 0 | 1 | 1 | 0 |

$(w^1)(w^2)\quad(w^3)(w^4)\quad(w^5)$

Figure 5: A non-uniform worldview that cannot be immediately coarsened. The state space has three binary dimensions (eight states). The worldview has two concrete states, $w^1$ and $w^4$, and three abstract states, $w^2$, $w^3$ and $w^5$, each abstract in a different dimension.

coarsened in any single dimension. Before any coarsening is possible, one of the states must first be refined, but if they all have low $\mathcal{P}$ they are not candidates for proximity-based refinement. Because of this, integration of a uniform abstraction method into coarsening would also not be as straightforward as for selecting the initial worldview or for refinement, unless the worldview is kept uniform. However, even with a non-uniform worldview it would not be difficult. For instance, the dimension could simply be removed only where possible rather than everywhere.

## 8. Results

We run the algorithm over a range of different domains to demonstrate our approach. The domains divide into two broad groups. The first group consists of the grid navigation domain only. This was the domain on which intuition was gathered and preliminary runs were done, however, so while problems in this domain show how well the approach performs, they cannot show its generality. The second group consists of domains from the literature, demonstrating how well our approach generalises.

### 8.1 Experimental Domains

We introduce our domains in this section. The first five problems are in the grid navigation domain, two already described in Section 2.1, shown in Figure 1, with three additional problems. The remaining domains are based on domains from the literature, in particular as used by Kim (2001) and Barry (2009).

As described in Section 2.1, problems in the grid navigation domain shown in Figure 1 all have x and y dimensions of size 10, three door dimensions (binary: open/closed) and a damage dimension (also binary). So far this is the 3Doors problem, while in the 3Keys problem, there are keys which the agent must pick up keys to open the corresponding doors. The three remaining problems are 1Key, shuttlebot and 10×10. The 1Key problem is similar to 3Keys, except that the agent is only capable of holding one key at a time, so that instead of three binary dimensions for the keys, there is one four-valued dimension indicating which key the agent holds (or none). The shuttlebot problem introduces a cyclic goal (with an extra 'loaded' dimension and with the damage dimension tri-valued)

(a) grid navigation domain

| Problem | keys in world | keys held at same time | note | dimensions | $|\mathcal{S}|$ |
|---|---|---|---|---|---|
| 3Doors | 0 | — | | 6 | 1 600 |
| 1Key | 3 | 1 | | 7 | 6 400 |
| 3Keys | 3 | 1, 2 or 3 | | 9 | 12 800 |
| shuttlebot | 0 | — | cyclic | 7 | 4 800 |
| 10×10 | 0 | — | tiled | 8 | 160 000 |

(b) robot$_4$-$k$ domain

| Problem | dimensions | $|\mathcal{S}|$ |
|---|---|---|
| robot$_4$-10 | 11 | 10 240 |
| robot$_4$-15 | 16 | 491 520 |
| robot$_4$-20 | 21 | 20 971 520 |
| robot$_4$-25 | 26 | 838 860 800 |

(c) factory domain

| Problem | dimensions | $|\mathcal{S}|$ |
|---|---|---|
| s-factory | 17 | 131 072 |
| s-factory1 | 21 | 2 097 152 |
| s-factory3 | 25 | 33 554 432 |

(d) tireworld domain

| Problem | locations | initial | goal | route length | dimensions | $|\mathcal{S}|$ |
|---|---|---|---|---|---|---|
| tire-small | 5 | n1 | n4 | 3 | 12 | 4 096 |
| tire-medium | 8 | n0 | n3 | 4 | 18 | 262 144 |
| tire-large | 19 | n12 | n3 | 1 | 40 | 1 099 511 627 776 |
| tire-large-n0 | 19 | n0 | n3 | 3 | 40 | 1 099 511 627 776 |

Table 3: Experimental domains and problems, each with the dimensionality of the state space and its size.

while the 10×10 variant increases the size of the problem by tiling the grid 10× in each direction (by having two extra dimensions, xx and yy, of size 10). Table 3(a) summarises the problems in this domain.

The next two domains are based on those of Kim (2001). Firstly, the robot$_4$-$k$ domain, which are based on Kim's (2001) ROBOT-$k$ domain but reducing the number of actions to four. The robot$_4$-$k$ domain problems consist of a cycle of $k$ rooms as shown in Figure 6, where each room has a light, analogous to the doors of the 3Doors problem in that they enable the agent to move. The four actions in our variant are to go forward, turn the light in the current room on or off, or do nothing. The original formulation allowed the agent any combination of toggling lights and going forward, for a total of $2^{k+1}$ actions, but we have reduced this as our approach is not intended to approximate in the action space. The goal is to move from the first room to the last. There are $k+1$ dimensions for a state space of $k2^k$ states, as listed in Table 3(b).

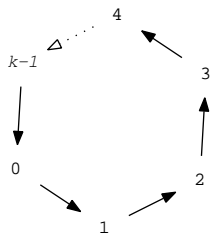

Figure 6: The $\text{robot}_4\text{-}k$ domain.

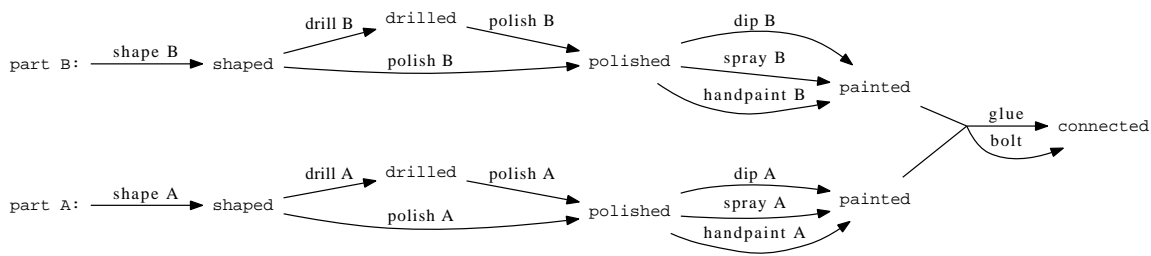

Figure 7: The `factory` domain.

Kim's (2001) `factory` domain[10] is a series of variants on a simple manufacturing problem, represented purely in predicates (that is, dimensions of size 2). The agent is to make a product from two parts which must be drilled, painted and so on and finally joined together. Figure 7 shows a very simplified diagram, omitting most of the interactions and options — for instance, to achieve the `painted` predicate, the agent may spray, dip or handpaint the object; to connect the two objects, it may use glue or a bolt (and only the latter requires that they be drilled); and so on. Unlike the other domains, partial rewards are available to the agent for achieving certain subgoals. The problems used here are listed in Table 3(c).

The final domain is the `tireworld` domain from the 2006 ICAPS IPC competition (Littman, Weissman, & Bonet, 2006) as used by Barry (2009). In this domain, a robotic car is trying to drive from point A to point B. The car has room to carry one spare tire and some locations have additional spare tires at them. At these locations, if the car is not carrying a spare, it can pick one up. If there are $n$ locations for the car, there will be $2n + 2$ binary dimensions in the problem, as follows: $n$ dimensions are used to represent the location of the car. The only valid states are states where only one location dimension is true, but this is not explicitly stated anywhere in the domain.[11] Another $n$ dimensions are used to represent which locations have a spare tire and which do not. The final two dimensions represent whether the car is carrying a spare and whether it has a flat tire.

---

10. This domain was previously used by Hoey, St.-Aubin, Hu, and Boutilier (1999) and is based on the builder domain of Dearden and Boutilier (1997) which was "adapted from standard job-shop scheduling problems used to test partial-order planners".
11. We touch on this aspect in our discussion in Section 9, but in any case include the domain without change in order to facilitate comparison with the literature.

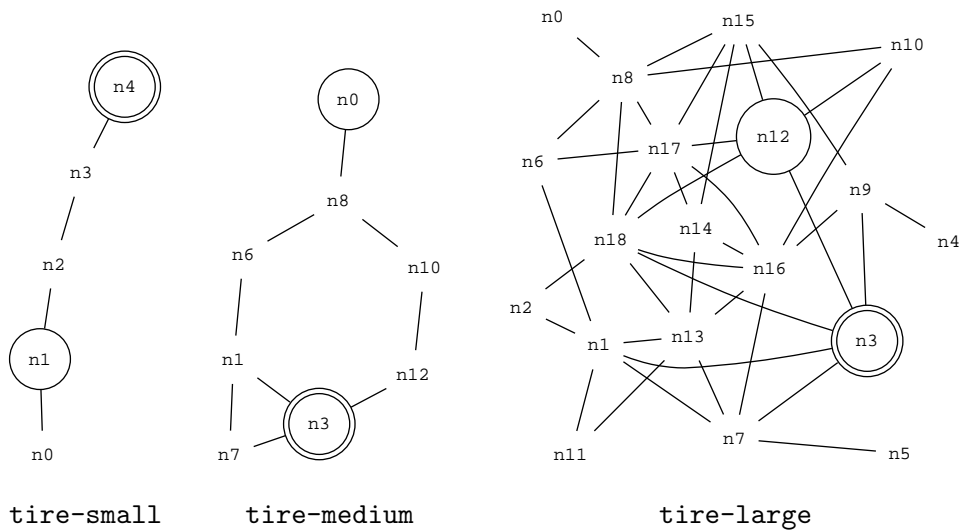

Figure 8: The tireworld domain problems, indicating the initial ($\odot$) and goal ($\circledcirc$) locations.

Barry (2009) uses two of the `tireworld` problems, labelled 'small' and 'large'. The small `tireworld` problem has 5 locations for 12 variables and 14 actions, while the large one has 19 locations for 40 variables and 100 actions. Curiously, in the large problem, there is a direct road between the initial and goal locations, so that it only takes a single action to solve the problem. This makes it difficult to assess whether Barry's method has, in fact, scaled up. In addition to these two, we created a medium-sized `tireworld`, with 8 locations and 18 variables, by removing locations from the large `tireworld` and moving the initial location to `n0`, further from the goal. These variants are listed in Table 3(d) and shown in Figure 8. The final variant, `tire-large-n0`, not shown, is identical to the large `tireworld` except the initial location is again moved to `n0`.

### 8.2 Direct Evaluation of Policies with Pre-cursor Deliberation

With the smaller problems such as `3Doors`, `1Key`, and `3Keys`, we can directly evaluate the approximate policies produced by the planner running with pre-cursor deliberation. These problems are small enough that we can use an exact algorithm to calculate the actual value function corresponding to these approximate policies. As noted in Section 2.6, this can be useful as it involves fewer potentially confounding variables, but it does not exploit the full potential of our approach.[12]

Table 4(a) shows such results for policy-based refinement only — that is, with the proximity-based methods (Algorithms 5, 6 and 7) disabled. For each problem, the table lists the size of the problem $|\mathcal{S}|$ and the value of the optimal solution at the initial state

---

12. Proximity-based coarsening (Algorithm 7) is primarily aimed at regions of state space that the agent has already traversed, but with pre-cursor deliberation there is no traversal. Coarsening would therefore be expected to bring limited benefit with pre-cursor deliberation and direct evaluation would not be meaningful to evaluate its performance. It is therefore only evaluated with recurrent deliberation.

$V^*(s^0)$, representing the costs and results for exact planning. These are followed by the size of the worldview $|\mathcal{W}|$ as an absolute number and as a percentage of $|\mathcal{S}|$, the planner's estimate of the value of its solution at the initial state $\hat{V}_{\hat{\pi}}(s^0)$, and the actual value of its solution at the initial state $V^*_{\hat{\pi}}(s^0)$. In the first half of each part of the table the discounting factor is $\gamma = 0.99999$, while in each second half it is $\gamma = 0.95$. Averages over 10 runs are shown and the planner is run to $\hat{\pi}_{1000}$. $\hat{\pi}_{1000}$ was chosen as an approximation to $\hat{\pi}_\infty$ — it is assumed that 1 000 phases is sufficient for the planner to converge. In practice, convergence generally took place much earlier. It is not detected, however, because of the overall assumption that the planner continues to plan forever, responding to any changing inputs, which makes convergence somewhat irrelevant.

| discounting factor $\gamma$ | problem | state space size $|\mathcal{S}|$ | optimal solution value $V^*(s^0)$ | worldview size $|\mathcal{W}|$ | relative worldview size $\frac{|\mathcal{W}|}{|\mathcal{S}|}$ | planner's estimate of solution value $\hat{V}_{\hat{\pi}}(s^0)$ | actual solution value $V^*_{\hat{\pi}}(s^0)$ |
|---|---|---|---|---|---|---|---|
| (a) policy-based refinement only | | | | | | | |
| 0.99999 | 3Doors | 1 600 | −27.50 | 226.4 | 14% | −22.50 | −27.50 |
| | 1Key | 6 400 | −79.47 | 326.8 | 5.1% | −37 512.20 | −100 000.00 |
| | 3Keys | 12 800 | −61.98 | 262.0 | 2.0% | −25 015.60 | −100 000.00 |
| 0.95 | 3Doors | 1 600 | −14.63 | 222.6 | 14% | −13.22 | −14.63 |
| | 1Key | 6 400 | −19.59 | 296.8 | 5.6% | −15.23 | −20.00 |
| | 3Keys | 12 800 | −18.99 | 245.7 | 1.9% | −14.56 | −20.00 |
| (b) proximity-based refinement only | | | | | | | |
| 0.99999 | 3Doors | 1 600 | −27.50 | 1 381.7 | 86% | −27.50 | −27.50 |
| | 1Key | 6 400 | −79.47 | 4 166.7 | 65% | −60 031.79 | −60 031.79 |
| | 3Keys | 12 800 | −61.98 | 4 262.8 | 33% | −70 018.59 | −70 018.59 |
| 0.95 | 3Doors | 1 600 | −14.63 | 1 363.2 | 85% | −14.63 | −14.63 |
| | 1Key | 6 400 | −19.59 | 2 828.2 | 44% | −19.92 | −19.92 |
| | 3Keys | 12 800 | −18.99 | 4 230.6 | 33% | −19.70 | −19.70 |
| (c) both policy- and proximity-based refinement | | | | | | | |
| 0.99999 | 3Doors | 1 600 | −27.50 | 1 361.7 | 85% | −27.50 | −27.50 |
| | 1Key | 6 400 | −79.47 | 4 635.5 | 72% | −20 063.57 | −20 063.57 |
| | 3Keys | 12 800 | −61.98 | 5 948.2 | 46% | −30 043.39 | −30 043.39 |
| 0.95 | 3Doors | 1 600 | −14.63 | 1 359.5 | 85% | −14.63 | −14.63 |
| | 1Key | 6 400 | −19.59 | 4 036.1 | 63% | −19.67 | −19.67 |
| | 3Keys | 12 800 | −18.99 | 3 748.8 | 29% | −20.00 | −20.00 |

Table 4: Results for direct evaluation of policies with pre-cursor deliberation with three different refinement methods, evaluated after 1 000 phases.

The results in part (a) of the table divide neatly into two types: without keys (3Doors problem), the planner succeeds in all ten runs, getting perfect policies for the given starting state. In the other two problems, 1Key and 3Keys, planning invariably fails. In these two problems, the agent must pick up a key while far from the door it opens, and this version of the planner simply cannot think ahead to that extent. For all three of these, the planner is somewhat optimistic, estimating a better value than it obtains and in some cases even better than optimum. For instance, in the 3Doors problem with $\gamma = 0.99999$, the planner's estimate of the value $\hat{V}_{\hat{\pi}}(s^0)$ is $-22.50$, which is better than both the true value and the optimum, $V^*_{\hat{\pi}}(s^0) = \hat{V}^*(s^0) = -27.50$. The fractional $|\mathcal{W}|$ in the table are due to its being averaged over ten runs. The final size of the worldview sometimes depends to some extent on the order in which the dimensions or the states are refined, and this order is randomised between the runs. For instance, in the 3Doors problem with $\gamma = 0.99999$, $\mathcal{W}$ has various sizes ranging from 220 to 231 states at the end of each of the ten runs, with an average of 226.4.

The results in part (a) are similar for the two values of $\gamma$. The main difference is that the smaller $\gamma$ leads to smaller numbers. For instance, the value indicating failure is $-100\,000$ when $\gamma = 0.99999$ but only $-20$ when $\gamma = 0.95$. The values tend to be in multiples of $\frac{1}{1-\gamma}$, and with the smaller value of $\gamma$ here, $\frac{1}{1-\gamma}$ is 20 rather than 100 000. In some cases, this smaller range can make differences less obvious: for instance, in the the estimated value column ($\hat{V}_{\hat{\pi}}(s^0)$), it is not very clear whether the numbers are approximations to $-1 \times \frac{1}{1-\gamma}$ (and a failure to reach the goal) or $0 \times \frac{1}{1-\gamma}$ minus a small number (representing success). While units represent once-off rewards or costs, multiples of $\frac{1}{1-\gamma}$ represent rewards or costs to be obtained in perpetuity. However, this is the expected and desired behaviour. A smaller $\gamma$ represents a disinterest in the distant future, so that a reward or cost in perpetuity is not much more important than a once-off reward or cost.

Table 4(b) shows the results of ten runs in pre-cursor mode to $\hat{\pi}_{1000}$ with proximity-based refinement only (no policy-based refinement and no coarsening) for each problem and $\gamma$. As can be seen, the 3Doors problem is solved optimally in all cases. This is not surprising, as it is not a complex problem.

The 1Key and 3Keys problems are more interesting. The figures in Table 4(b) arise as the average of about $\frac{1}{3}$ successful runs, which have values close to or equal to the optimal values $V^*(s^0)$, and about $\frac{2}{3}$ unsuccessful runs which have values of $-100\,000$. For $\gamma = 0.99999$, the planner found a successful policy in 4 of the 10 runs in the 1Key problem and 3 times out of 10 in the 3Keys problem. Similarly for the $\gamma = 0.95$ case (2 times and 3 times, respectively), but since the optimal path is quite long compared with $\frac{1}{1-\gamma}$, so that success means a reward of $-19.59$ (or $-18.99$) while failure is punished by $-20$, the effect is more difficult to discern.

Table 4(c) shows the results for proximity-based refinement and policy-based refinement combined (no coarsening). Naturally, the 3Doors problem for which either refinement method alone already obtained the optimal policy shows no improvement. Only the worldview size $|\mathcal{W}|$ differs slightly from Table 4(b), as policy-based refinement is sometimes more directed, so the $|\mathcal{W}|$ tend to be slightly smaller than with the more exploratory proximity-based refinement alone, but larger than with policy-based refinement alone.

The other two problems, `1Key` and `3Keys`, show improvement compared with either of the refinement methods alone. They are now solved in $\frac{3}{4}$ of the runs for $\gamma = 0.99999$ — the values $-20\,063.57$ and $-30\,043.39$ represent averages between 2 and 3 unsuccessful runs and 8 and 7 successful ones, compared with 4 and 3 successful runs for proximity-based refinement only and no successful runs for policy-based refinement only. For $\gamma = 0.95$, the `1Key` problem is again solved on 8 of the runs, but due to the discounting and the length of the path, the goal is very near the horizon. Again, with success meaning a reward of $-19.59$ while a failure receives $-20$, the distinction is not very great. The `3Keys` problem with $\gamma = 0.95$ did not find a solution at all with these parameters, so it receives a uniform $-20$ for each of its runs, for a suboptimality of about one unit.

The behaviour during these runs is generally quite straightforward. Typically, after initially calculating that the agent cannot reach the goal with the initial worldview, the worldview size gradually increases, then plateaus — there is no coarsening here, nor movement of the agent, so no other behaviour is really possible. In the successful runs, the planner plans a route to the goal at some point during the increase, when the worldview becomes sufficient, and $\hat{V}_{\hat{\pi}}(s^0)$ quickly reaches its final value. Rarely, the $\hat{V}_{\hat{\pi}}(s^0)$ may oscillate once or twice first. We omit the graphs here, but they are presented by Baum (2006).

### 8.3 Evaluation by Simulation with Recurrent Deliberation

In larger problems, performance can be evaluated by simulation, running the agent in a simulated world and observing the reward that it collects. In such problems, direct evaluation is not possible because calculating the actual value function using an exact algorithm is no longer tractable. Simulation with recurrent delibertion is also the context in which coarsening can be evaluated. For comparison, this section presents results for the `3Keys` problem evaluated by simulation, both without and with coarsening.

Figure 9 shows a representative sample of the results for simulation on the `3Keys` problem with both refinement methods and no coarsening, which is the same combination of options as shown in Table 4(c) in the previous section, evaluated by simulation rather than directly. Each small graph shows a different individual run. As can be seen, the agent behaves reasonably when working in the recurrent planning mode against the simulation.

The left vertical axes on the graphs represent reward $R$, plotted with thick red lines. In run 1 of Figure 9, for instance, the agent starts off receiving a reward of $-1$ for each step, meaning 'not at goal, no damage' in this domain. From about 180 onwards, it receives a reward of 0 per step, meaning 'at goal, no damage'. The right vertical axes are the worldview size $|\mathcal{W}|$, with thin blue lines. They are shown as details throughout this section, that is, scaled to the actual worldview sizes rather than the full $1$–$|\mathcal{S}|$ ranges. Taking run 1 in Figure 9 again, we see that $|\mathcal{W}|$ grows relatively quickly until about 80, then continues to grow slowly and eventually levels out a little below $5\,000$. The full state space, for comparison, is $12\,800$. In run 2, the agent received reward similarly, but the state space grew longer, eventually levelling out somewhat above 7000. In runs 3 and 4, the agent failed to reach the goal and continued receiving the reward of $-1$ throughout. In run 3, the worldview size levelled out a little over 3000, while in run 4 it steadily grew to about 5000.

The horizontal axes are simulated world time, corresponding to the discrete time-steps of the MDP. There are two other time-scales in the simulation: wall clock time, indicating the passage of real time, and the number of phases the planner has performed. The simulation is configured to take 1 time step per 10s of wall clock time. The number of phases is not controlled and is simply given by the speed of the (1.5GHz Intel®) CPU and an implementation coded for flexibility rather than efficiency. Ideally, the agent should gradually move in the general direction of the goal during planning, as this simplifies the problem, but not so fast that the agent runs too far ahead of the abstraction in the planner.

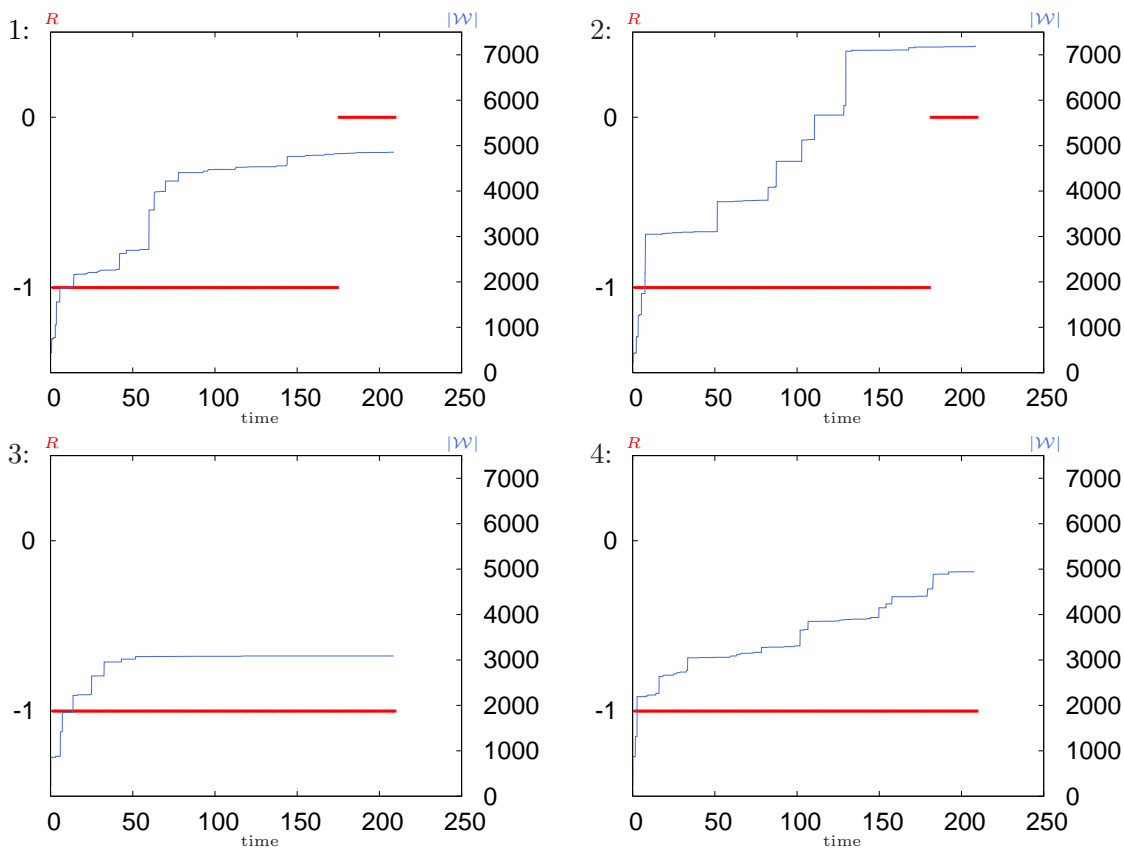

Figure 9: Simulation results, 3Keys problem, policy-based and proximity-based refinement, no coarsening (four runs). Reward (left axes, thick red lines) and worldview size $|\mathcal{W}|$ (right axes, thin blue lines; detail) against world time (horizontal axes).

The planner algorithm does not terminate, since the planner is assumed to keep planning (and the agent to keep acting) indefinitely. In goal-oriented domains, such as most of the examples in this paper, one might consider achieving the goal to be such a termination condition, but (a) the example domains assume that the agent will continue with the same goal as a goal of maintenance, albeit trivial, (b) it does not apply at all to non-goal-oriented domains and (c) even in goal-oriented domains it is not clear how to apply the condition in the case where the agent fails to reach the goal. In the simulation, therefore, runs were either terminated manually, when they succeeded or when it appeared that no further progress was likely to be made, or run for a fixed number of world time steps, selected based on the manually-terminated runs with some allowance for variation.

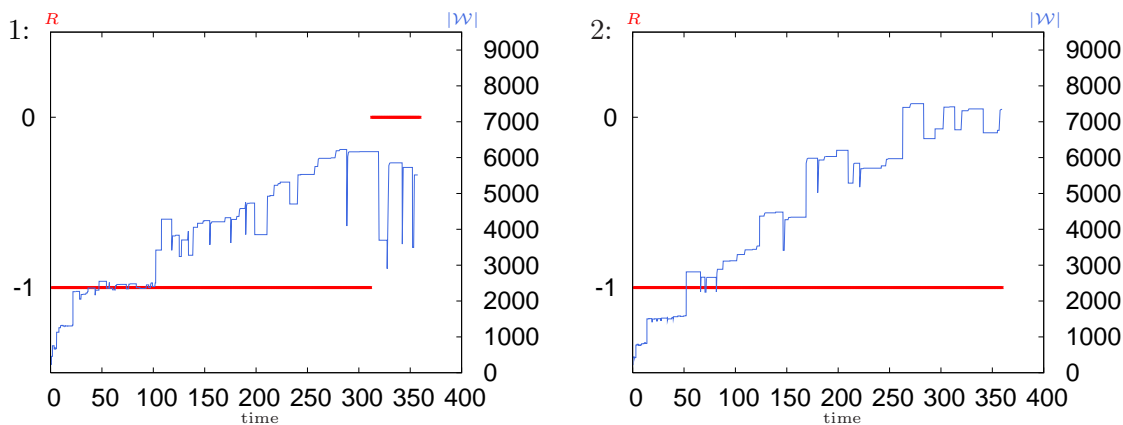

Figure 10: Simulation results illustrating the effect of coarsening on worldview size, `3Keys` problem, policy-based refinement, proximity-based refinement and coarsening (two runs).

Because there is no coarsening in Figure 9, the worldview sizes are monotonic increasing. Different runs refined differently — both the domain and the algorithm are stochastic. At the beginning of planning, the agent is receiving a reward of $-1$ per step, because it is not yet at the goal. As the worldview size increases, the planner eventually finds a policy which leads to the goal in runs 1 and 2, as can be seen by the better reward 0 obtained in those runs. There is a simple relationship between worldview size and performance: runs which worked with a worldview of about 5 000 or larger generally succeeded, those with smaller worldviews generally did not. On the vast majority of the runs ($> 90\%$), the agent reached the goal.

When coarsening (Algorithm 7) is activated, compared to the situation when it is turned off, the reward gathered by the agent declines slightly, but it still reaches the goal on the vast majority of runs ($> 90\%$). Figure 10 shows two of the runs, one successful and one unsuccessful, for the `3Keys` problem with proximity-based coarsening as well as the two refinement methods, in contrast with Figure 9 where only the two refinement methods are used. Note the effect of the interleaving of the refinement and coarsening: the worldview size $|\mathcal{W}|$ (thin blue line) is no longer monotonic, instead being alternately increased and decreased, which shows up as a jagged line on the graph. Slightly fewer of the runs reach the goal. Some decline in solution quality is expected, however, since the goal of coarsening is to reduce the size of the worldview.

For completeness, we have also tested the agent with both proximity-based methods (Algorithms 6 and 7) active but with the policy-based refinement (Algorithm 4) deactivated. In this configuration, the agent collects no reward — it generally takes some steps toward the goal, but without the more directed policy-based refinement, the largely exploratory proximity-based methods do not discover the keys and consequently cannot reach the goal.

### 8.4 The Effect of the Discounting Factor

The `shuttlebot` problem is similar to the `3Doors` problem but requires the agent to move back and forth between two locations repeatedly. It is interesting because in preliminary runs in pre-cursor mode it was not solved at all for $\gamma = 0.99999$ while being solved optimally for $\gamma = 0.95$. It was considered whether the $\gamma = 0.99999$ case might behave better under simulation, if the agent took advantage of the possibility of planning only for the nearest reward and then replanning once that reward is obtained. After all, the agent could function well even if none of the policies were a good solution by itself. However, as illustrated in Figure 11, the agent's behaviour was very similar to the pre-cursor case: for $\gamma = 0.99999$, (a) refinement only, run 1, and (b) with coarsening, run 2, it would pick up the reward immediately adjacent to $s^0$, but no more than that. Again, setting the planner's discounting factor $\gamma$ to 0.95, (c) with coarsening, runs 3 and 4, provided much better performance.[13] Note again the effect of the balance of refinement and coarsening in (b) and (c): the worldview size $|\mathcal{W}|$ is nice and steady throughout the runs (though admittedly at a fair fraction of $|\mathcal{S}| = 4\,800$).

### 8.5 Initial Worldview

For some of the problems, the standard initial worldviews are too large for the planner. Even modified, smaller initial worldviews obtained by only enabling the nexus step of Algorithm 3 and disabling the reward step are too large. Disabling both the reward and the nexus steps results in a singleton initial worldview, $\mathcal{W} = \{\mathcal{S}\}$, which treats the entire state space $\mathcal{S}$ as a single (very) abstract worldview state. Unfortunately, this means that the planner starts with very little in the way of hints as to the direction in which to refine and, at least initially, no other information on which to base this crucial decision. The upshot is that it collects no reward. In some cases it remains at the initial state $s^0$, in others moves around the state space — sometimes for some distance, other times in a small loop — but it does not reach the goal or any of the subgoals.[14]

This is the situation with the factory domain problems of Kim (2001), and in fact the agent collected no reward during any of the simulated runs, even though in quite a few runs substantial actions were taken. A similar result occurs for the `10x10` problem (in our grid navigation domain). No reward was obtained by the agent in this problem, because the standard initial worldview is somewhat too large for the planner and, again, the singleton initial worldview does badly. At best, on some runs, the agent took a few limited steps in the general direction of the goal.

An interesting case is the tireworld domain. Again, `tire-large` is too large with the standard initial worldview and fails to obtain a solution with the reward step of the initial worldview only. However, with a manually-chosen initial worldview that refines the locations along the path from the start state to the goal before planning begins, the planner solves not only `tire-large`, but also `tire-large-n0` in 40% of the runs (in one of them after less than one minute, although that is atypical).

---

13. The rewards appear as two horizontal lines in runs 3 and 4, one solid and one broken, because the task is cyclic, and the agent collects a reward of 1 twice in each cycle and a reward of 0 on all the other steps.
14. Further details of these unsuccessful runs, including the $|\mathcal{W}|$ behaviour, are given by Baum (2006).

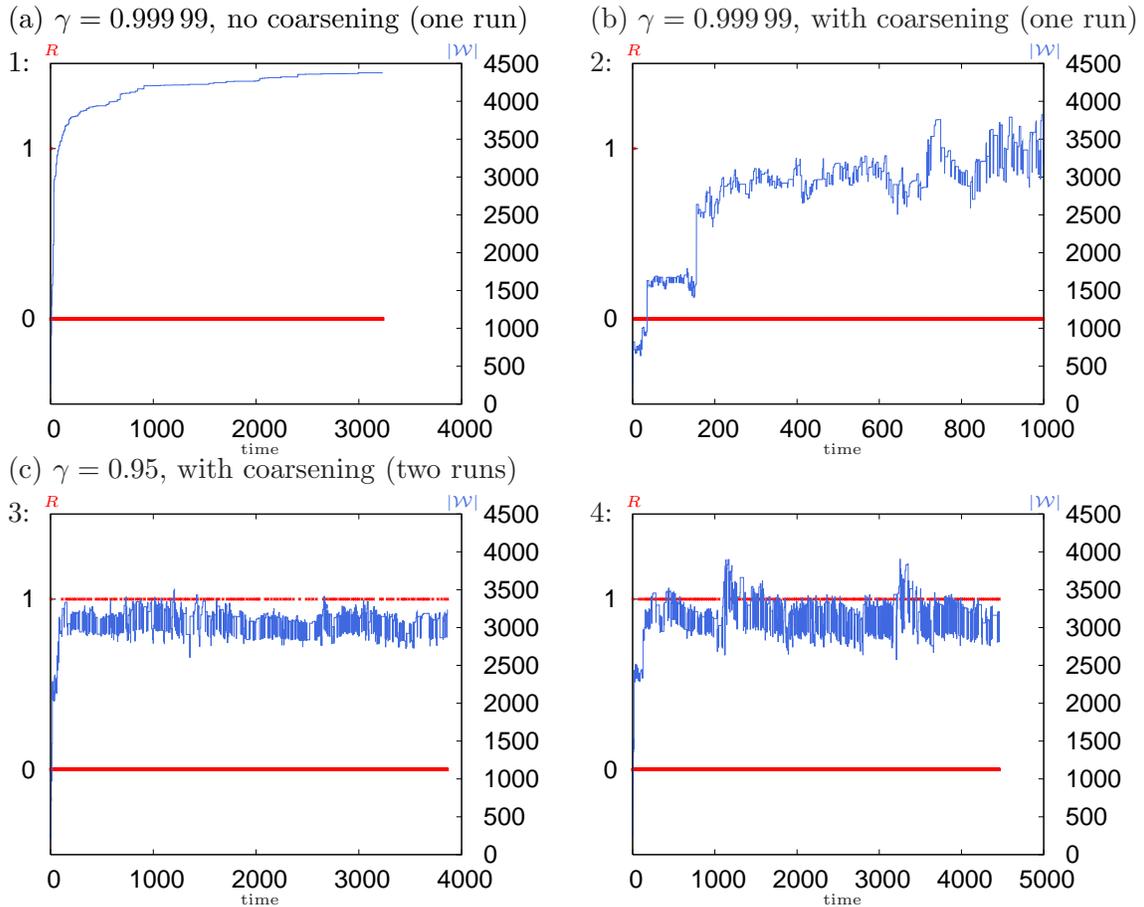

Figure 11: Simulation results illustrating the effect of the discounting factor, `shuttlebot` problem, policy-based and proximity-based refinement.

### 8.6 Worldview Size and Quality

Finally, we consider the effect of worldview size and quality on the $\texttt{robot}_4$ domain, where the agent moves through a series of rooms with lights. This domain is an excellent example where the simulated agent works well. In all runs of the $\texttt{robot}_4$-10 and $\texttt{robot}_4$-15 problems the agent thought for a small amount of time, then quickly moved to the goal and stayed there, with only very small worldviews, as can be seen in Figure 12 for the $\texttt{robot}_4$-15 problem. Four representative runs are shown, two with the two refinement methods only (runs 1 and 2) and two with all three methods (runs 3 and 4). In all four runs, the worldview sizes $|\mathcal{W}|$ are reasonable — consider that the full state space contains almost half a million states, so that a 1 000-state worldview represents just a fifth of a percent. Despite this small worldview size, however, the planner is effective. After only a few dozen phases, the agent has reached the goal. The planner works well for $\texttt{robot}_4$-10 and $\texttt{robot}_4$-15.

For comparison, Kim's (2001) largest ROBOT-$k$ problem is ROBOT-16, though since ROBOT-16 has $2^{16+1} = 131\,072$ actions while our $\texttt{robot}_4$-$k$ domain problems have 4, a direct comparison would not be valid. On the other hand, our values of $k$ (10, 15 and so on) are not necessarily powers of 2, since, unlike that of Kim, our domain specification always considers room numbers atomic rather than binary numbers, so there is no particular advantage to powers of 2.

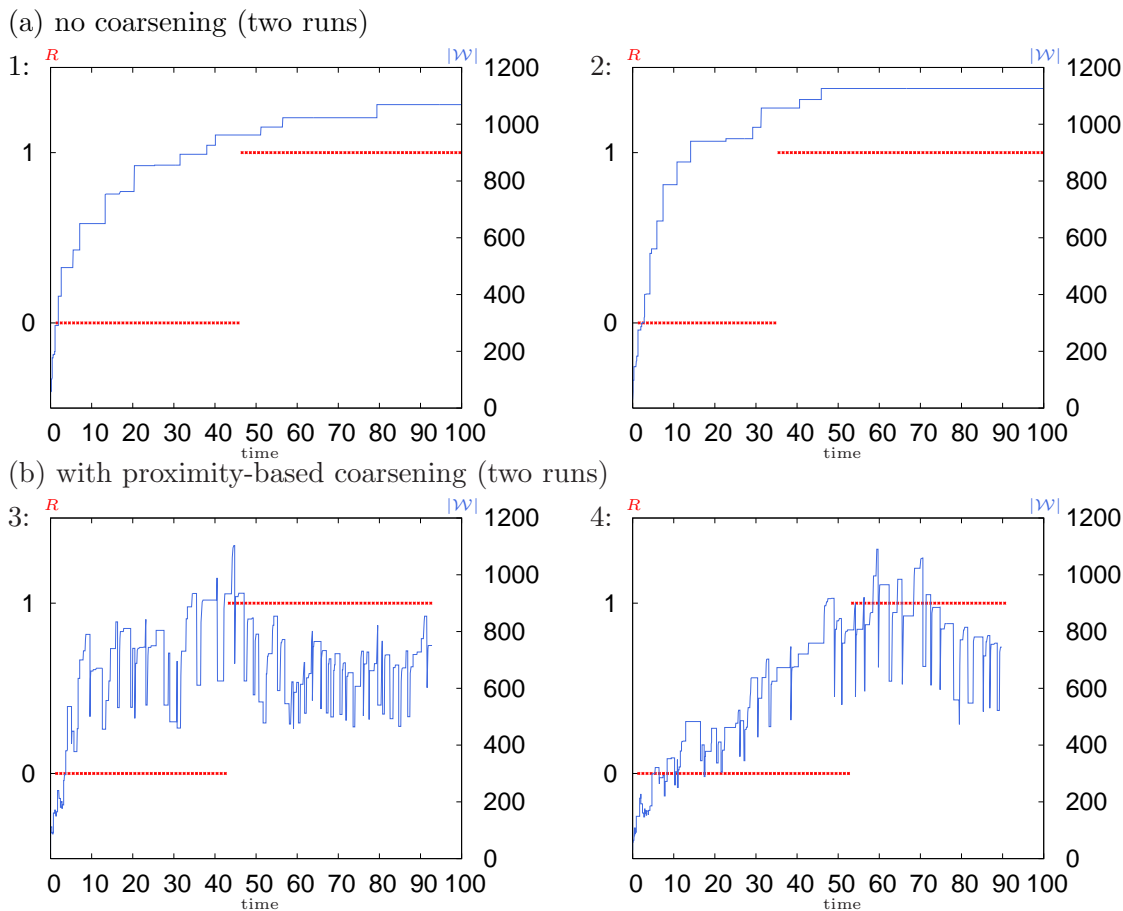

Figure 12: Simulation results, robot$_4$-15 problem, $\gamma = 0.99999$, policy-based and proximity-based refinement, with and without proximity-based coarsening.

The results for the robot$_4$-20 problem are beginning to be more interesting than those for robot$_4$-10 and robot$_4$-15. In Figure 13(a), showing two of the runs with no coarsening (runs 1 and 2), the agent succeeds reasonably promptly and with reasonable worldview sizes. However, as illustrated in Figure 13(b), when coarsening is active the planner fails to reach the goal on some of the runs (about 40%, for example, run 3) and succeeds on others (about 60%, for example, run 4). The state space contains almost 21 million states, so the successful worldviews in Figure 13 are of the order of 0.01% of the full state space size.

Figure 14 shows four representative runs of the robot$_4$-25 problem, again (a) two without coarsening (runs 1 and 2) and (b) two with all three methods (runs 3 and 4). This problem has a state space of $25 \times 2^{25} \approx 839$ million states, and the effect noted for robot$_4$-20 is much more pronounced here: without coarsening, the planner tends to much larger worldviews[15] These large worldviews then cause the planner to run slowly. As noted in Section 8.3 above, the horizontal axes are world time, not planning time. The relation between the two varies quite significantly between the runs — from more than two policies per time step with the smaller worldviews to less than one in ten time steps when the worldviews grew large.

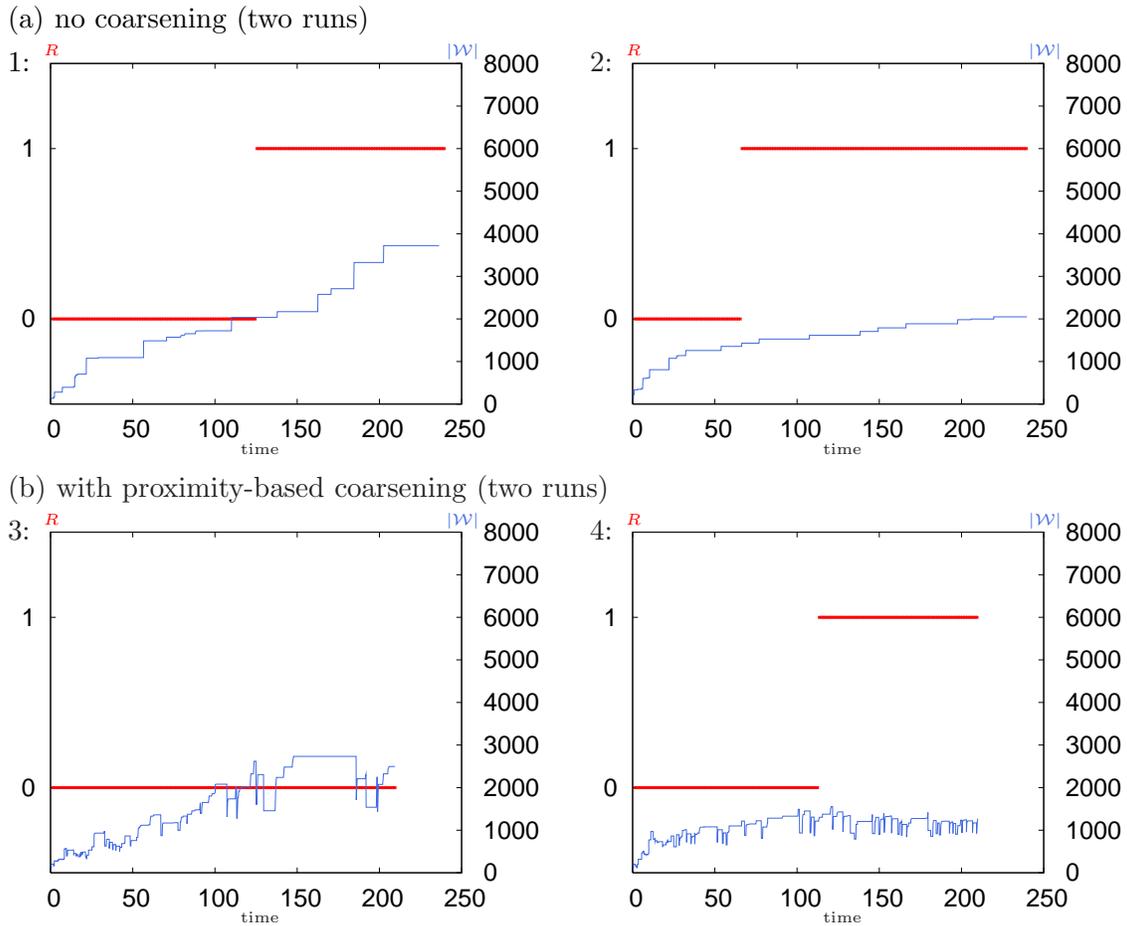

Figure 13: Simulation results, robot$_4$-20 problem, $\gamma = 0.999\,99$, policy-based and proximity-based refinement, with and without proximity-based coarsening.

As far as reaching the goal is concerned, the two cases are similar. Again, the successful runs are those which maintain a reasonably-sized worldview, such as runs 2 and 4. Runs where the worldview size grows big invariably fail (runs 1 and 3). The difference is that this time, the smallest successful worldview, run 4 in Figure 14, used around 2 000 well-chosen worldview states, which is just $0.000\,24\%$ of the full state space. If the worldview grows beyond a miniscule fraction of the state space — and even the 19 243-state worldview of run 1 in Figure 14 is only $0.0023\%$ — the planner will stall and no further progress will be possible. Even in this challenging environment, the agent reaches the goal in almost half the runs.

---

15. Note that run 1 is plotted with a different scale on the worldview size $|\mathcal{W}|$ axis compared to runs 2, 3 and 4. This makes the details of their behaviour easier to see, but makes the size of run 1 less obvious.

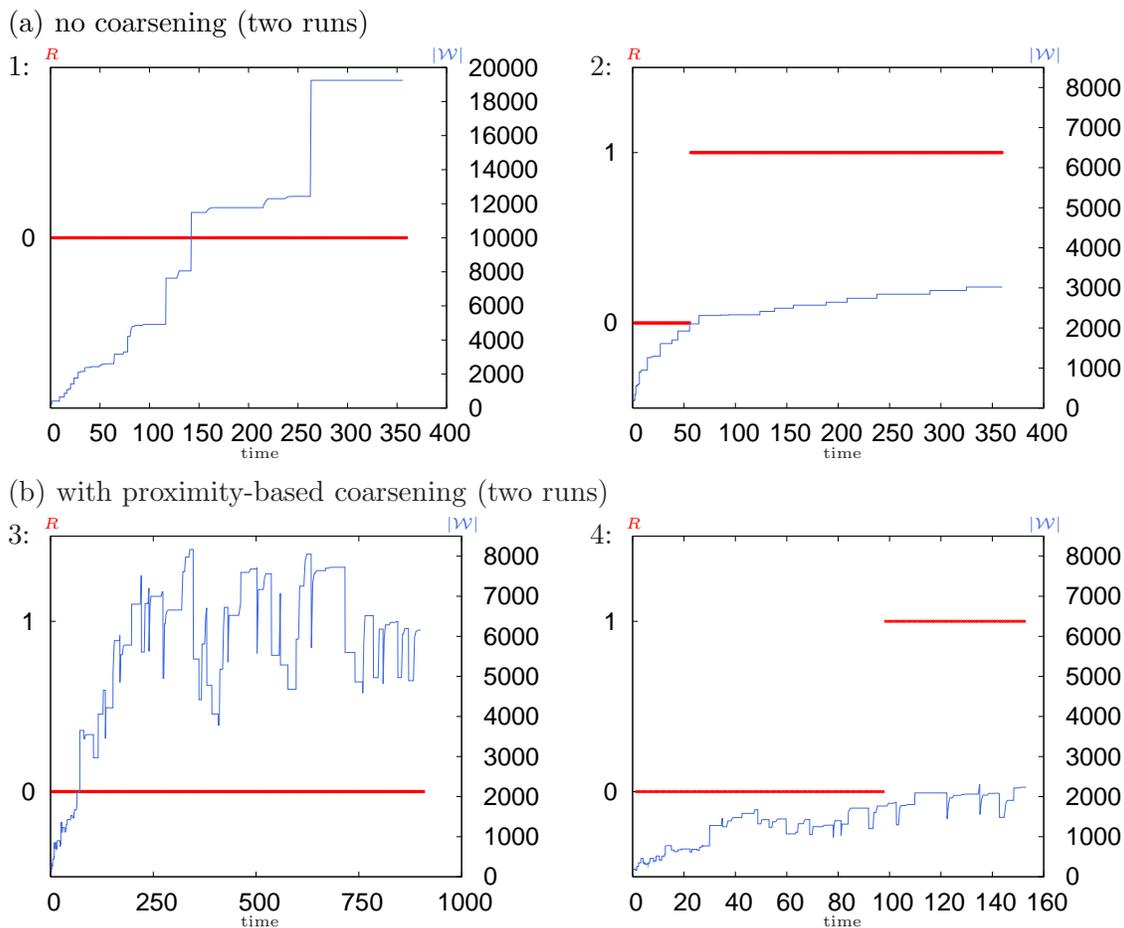

Figure 14: Simulation results, `robot_4`-25 problem, $\gamma = 0.99999$, policy-based and proximity-based refinement, with and without proximity-based coarsening. Note the different scales on the worldview size $|\mathcal{W}|$ axis for run (a)1.

## 9. Discussion

Section 8 presented results across a range of experimental domains showing that our method successfully finds solutions to the planning problem using much less than the full state space, as well as some of its limitations. In this section we discuss these results and analyse the features of domains which our method can exploit and those which give it difficulty.

On smaller problems, we could directly evaluate the policies produced by our method in pre-cursor mode, allowing us to better isolate the behaviour of the planner. Without the proximity-based methods, the worldviews were quite small and the planner could only solve the `3Doors` problem. However, even this is better than a uniform abstraction, which could do very little here. Even an oracle could at best remove one door and its key in `3Keys` or two doors in `3Doors`, giving a 25% relative worldview size — however, if the planner then used a uniform distribution over the removed dimension, the agent would fail anyway, since opening the doors that were left in the worldview would be harder than hoping for the best on the assumed-50%-open door which is actually closed. To succeed, it would also have to deduce that the abstracted doors should be considered closed, a considerable feat. This is a function of the sample domains. In other circumstances, uniform abstraction could be very effective, either as a pre-processing step to our approach or integrated with our $\mathcal{W}$ selection methods.

We expected that turning on proximity-based refinement (Algorithms 5 and 6) would lead to larger worldviews. In general, one would expect larger worldviews to yield better solutions and smaller worldviews to yield worse solutions, but for lower computational cost. This means that proximity-based refinement should, in general, improve the solution quality. The results corresponded to this expectation. The worldviews were indeed larger and the solution quality was higher.

In larger problems, performance could only be evaluated by simulation, running the agent in a simulated world and observing the reward that it collects. In such problems, direct evaluation was not possible because calculating the actual value function using an exact algorithm was no longer tractable. Section 8.3 therefore presented results for the `3Keys` problem for comparison with those obtained by direct evaluation of policies with pre-cursor deliberation discussed above. As can be seen, the results correspond, cross-confirming the evaluation methods.

In addition, simulation with recurrent delibertion is the context in which coarsening could be evaluated. When proximity-based coarsening was activated, compared to the situation when it is turned off, the reward gathered by the agent declines slightly. This impression of worse performance is somewhat misleading. It is due to the fact that this first comparison takes place on one of the small problems, which the planner was able to solve without any coarsening, and in fact without any abstraction at all. Thus, the disadvantages (lower reward collected) are much more apparent than the advantages (lower computational cost).

On larger problems, where working without abstractions was not an option, the balance was reversed. In fact, and somewhat counterintuitively, in some of the larger problems with coarsening active, the successful runs had smaller worldviews than the unsuccessful runs. Clearly, it is not the size of the worldview that determines success, but its quality. A good worldview enabled efficient calculation of policies that progress toward the goal while remaining small. A poor worldview simply grew larger. In the smaller problems, a growing worldview may have eventually covered most of the state space in detail, thus masking the effect. The planner would find a good policy effectively without any real approximation. In larger problems, where finding a good policy without approximation was not feasible, a similarly growing worldview simply slowed the planner down until no further progress was made. In this situation, the worldview-reducing action of proximity-based coarsening became crucial, ensuring that at the least the worldview remained tractably small and thereby enabled the planner to deal with the problem.

As can be seen from the results, coarsening was successful in this task some of the time. When the agent paused to replan part-way to the goal, it reduces the size of the worldview to keep it more relevant to the changing circumstances. In other runs, however, the worldview size grew beyond the capabilities of the planner. In some cases, such as in the `10x10` problem, it settled at a higher balance. In others, it appears to have simply continued growing. In the latter case, it would not appear to be a simple question of tuning the parameters: when it did find balance, it was at an appropriate worldview size. There appears to be some other factor, some other quality of the worldview determining the success or failure of those runs — whether they find balance and reach the goal or grow too big and fail.

A number of problems were solved poorly or not at all due to our initial abstraction selection algorithm (Algorithm 3). On these problems, the algorithm produced either a large worldview that exceeded available memory (either immediately or very shortly afterwards), so that planning was not possible, or a small worldview in which planning was ineffective. It could not be set to produce a medium-sized worldview, none of the four combinations of options produced one. In some problems, only the singleton worldview was possible — that is, both steps of initial abstraction selection disabled, resulting in a worldview aggregating all states into a single, maximally abstract worldview state — leading typically to no reward being collected by the agent. At best, it would take a few actions in the general direction of the goal(s). This is considerably worse than previous work. For instance, Kim (2001) obtains approximate solutions for the problems and for their larger variants, as do others who use this domain or a variant, including Hoey et al. (1999) and their originators, Dearden and Boutilier (1997).

It was the necessity of using a singleton initial worldview which understandably greatly hurt the performance. An infelicity at the worldview initialisation stage could impair the entire planning process, since the worldview was never completely discarded by the planner. The worldview-improvement algorithms could have made up for some amount of weakness in the initial worldview, but a singleton worldview was a poor starting point indeed. This is similar to an observation made by Dean et al. (1995) in their work using a reduced envelope of states (that is, a subset of the state space). Their high level algorithms, like those here, work regardless of the initial envelope (worldview). In practice, however, it is better for the initial envelope to be chosen with some intelligence, for instance to contain at least a possible path to the goal (for goal-oriented domains). They find this path using a simple depth first search, and while this itself is not directly applicable to worldviews with their gradations of abstraction, the overall concept remains: a reasonable initial worldview is crucial.

The tireworld results confirm this. Here the initial worldview with the reward step enabled was small, since the domain only rewards a single dimension, and planning was ineffective. Once the planner was given a better initial worldview — one which it could have plausibly calculated — it became quite effective, even in the modified `tire-large-n0` where the initial state was deliberately moved further from the goal. It seems, then, that while our basic approach is general, our worldview selection and modification methods are less so. This is good — worldview selection is the less fundamental apect of our approach and can be easily supplemented by additional methods or even just tuning. In domains like tireworld, it seems that a modified predicate solver that can generate plausible trajectories from the current state to a goal would do well as part of worldview selection.

It is interesting to compare this with the results of Sanner and Boutilier (2009) for tireworld, which they only describe in passing as "extremely poorly approximated" before going on to manually tweak the domain for their planner, adding the information that the locations are mutually exclusive. This makes planning much easier and largely invalidates any comparison between approaches.[16] It is a fair question, however, to what extent this is a weakness of their planner and to what extent it is an artefact of the domain. Its combination of representation and narrative seems rather unfortunate, as the narrative with its obvious 1-of-$n$ intuition will tend to obscure the real features (and the real applicability) of the propositional representation and hinder rather than help intuition. This would not occur with a tighter fit between representation and narrative.

Others, of course, solve the tireworld domain well. Some, such as Barry, Kaelbling, and Lozano-Prez (2010), generate the full policy, while others take advantage of the initial state, as our planner would do. It is difficult to know to what extent their planners are adapted to the domain and to what extent they are flexible. It seems that only in recent years has it become common for planners to be tested on domains to which the researchers do not have access during development, as with some of the ICAPS IPC domains, rather than being to a greater or lesser degree hand-tuned, usually unconsciously, to the particulars of one or another domain, as ours was undoubtedly unconsciously tuned to the grid navigation domain.

An interesting side point is provided by the `shuttlebot` problem, which was not solved at all for $\gamma = 0.999\,99$ (other than collecting the trivial reward immediately adjacent to $s^0$) while being solved optimally for $\gamma = 0.95$. Since the simulator itself does not have any intrinsic discounting factor — it reports each reward as it is collected — one can see that even though the planner was working with a discounting factor $\gamma = 0.95$, it provided a better solution to the $\gamma = 0.999\,99$ case than when it worked with $\gamma = 0.999\,99$ in the first place.

In some ways this better behaviour with the smaller planner discounting factor is reasonable, because while the agent's horizon is represented by the world discounting factor, the horizon of any particular policy — and therefore the planner — is effectively much shorter, as the policy will be supplanted by a new one relatively soon. Thus, it may be useful on occasion to set the planner's discounting factor $\gamma$ lower than the true world discounting factor in order to facilitate planning. However, this may lead to suboptimal, "short-sighted" policies.

## 10. Conclusions

The theory of Markov decision processes provides algorithms for optimal planning. However, in larger domains these algorithms are intractable and approximate solutions are necessary. Where the state space is expressed in terms of dimensions, its size and the resulting computational cost is exponential in the number of dimensions. Fortunately, this also results in a structured state-space where effective approximations are possible.

---

16. Similarly, Kolobov, Mausam, and Weld (2008) report results on a variant of tireworld rather than on tireworld itself, without providing an explanation.

Our approach is based on selectively ignoring some of the dimensions in some parts of the state space in order to obtain an approximate solutions at a lower computational cost. This non-uniform abstraction is dynamically adjusted as planning and (in on-line situations) execution progress, and different dimensions may be ignored in different parts of the state space. This has strong implications, since the resulting approximation is no longer Markovian. However, the approach is both intuitive and practical. It is the synthesis of two existing approaches: the structure-based approximation of uniform abstraction and the dynamic locality-based approximation of envelope methods. Like the envelope methods, it can be limited by its reliance on the initial worldview (or envelope): if that is poor, it will tend to perform poorly overall. Our approach subsumes uniform abstraction completely — it can be treated as a special case of our more general method.

This paper extends the preliminary work of Baum and Nicholson (1998) by modifying the worldview based on the proximity measure, both enlarging and reducing its size, and by evaluating the behaviour against a simulation. This allows us to test the approach on larger problems, but more importantly demonstrates both the full strength of the approach and its limits in terms of the domain features that it can exploit and those that it can exploit only with adjustment or not at all. The abstraction becomes truly dynamic, reacting to changes in the agent's current state and enabling planning to be tailored to the agent's situation as it changes. As shown by the qualitative and quantitative results presented both here and by Baum (2006), the approach can be effective and efficient in calculating approximate policies to guide the agent in the simulated worlds.

### 10.1 Future Work

One possible direction for future research would be to find worldview initialisation and modification methods that result in smaller yet still useful worldviews, probably domain-specific, to extend the method to further domains, either larger or with different features. For example, the factory and tireworld domains are goal-oriented and based on predicates, and a worldview selection or modification method based on a predicate-oriented solver could find possible paths to the goal and ensure that relevant preconditions are concrete along that path.

Interestingly, in the `10x10` problem, while the proximity-based methods did not keep the worldview size small, they did seem to find a balance at a larger but stil moderate size. Thus another possibility might be to tune the proximity-based methods or develop self-tuning variants.

At a number of points, for instance phase selection, our algorithm uses stochastic choice as a default. This could be replaced by heuristics, learning, or other more directed methods.

One could adapt our method to work with other types of MDPs, such as undiscounted or finite-horizon ones, or combine it with other approaches that approximate different aspects of the domains and the planning problem, as described in Section 2.5. For example, as mentioned in that section, Gardiol and Kaelbling (2004, 2008) combine hierarchical state space abstraction somewhat similar to ours with the envelope work of Dean et al. (1995). Many other combinations would likely be fruitful for planning in domains which have features relevant to multiple methods. Similarly, additional refinement or coarsening methods could be added, for instance as one based on the after-the-fact refinement criterion with roll-back of Reyes et al. (2009).

On a more theoretical side, one could look for situations in which optimality can be guaranteed, as Hansen and Zilberstein (2001) do with their LAO* algorithm for the work of Dean et al. (1995), observing that if an admissible heuristic is used to evaluate fringe states, rather than a pragmatically chosen $V(\text{OUT})$, the algorithm can be related to heuristic search and acquires a stopping criterion with guaranteed optimality (or $\varepsilon$-optimality). Perhaps a similar condition could be developed for our approach, with a rather different heuristic.

There are two basic directions in which this work can be further extended in a more fundamental way, relaxing one of the MDP assumptions, perfect observability or knowledge of the transition probabilities. A Partially Observable Markov Decision Process (POMDP) gives the agent an *observation* instead of the current state, with the observation partly random and partly determined by the preceding action and the current state. While the optimal solution is known in principle, it is quite computationally expensive, since it transforms the POMDP into a larger, continuous, many-dimensional MDP on the agent's beliefs. As such, the non-uniform abstraction approach could be applied in two different ways: either to the original POMDP, a fairly direct translation, or to the transformed MDP. The other extension would be to apply the technique when the agent has to learn the transition probabilities. In particular, the application of our technique to exploration[17] would be very interesting — the agent would have to somehow learn about distinctions within single abstract states, so as to distinguish which of them should be refined and which should remain abstract.

---

17. While the learning problem could also be transformed into an MDP on the agent's beliefs or experiences, it would be computationally prohibitive. The standard approaches instead explicitly distinguish 'exploration', where the agent learns about its domain (but ignores goals) and 'exploitation', where it achieves goals (but ignores opportunities to learn).